\documentclass{article}

\usepackage{PRIMEarxiv}
\usepackage{placeins}  % in preamble
\usepackage[utf8]{inputenc} % allow utf-8 input
\usepackage{float} % in preamble
\usepackage[T1]{fontenc}    % use 8-bit T1 fonts
\usepackage{hyperref}       % hyperlinks
\usepackage{url}            % simple URL typesetting
\usepackage{booktabs}       % professional-quality tables
\usepackage{amsfonts}       % blackboard math symbols
\usepackage{nicefrac}       % compact symbols for 1/2, etc.
\usepackage{microtype}      % microtypography
\usepackage{lipsum}
\usepackage{fancyhdr}       % header
\usepackage{graphicx}       % graphics
\graphicspath{{media/}}     % organize your images and other figures under media/ folder
\usepackage{multirow}
\usepackage{gensymb}
\usepackage{amsmath}       % for math symbols
\usepackage{algorithm}     % for algorithm float environment
\usepackage{algpseudocode} % for algorithmic commands (like \State, \For, etc.)
\usepackage{float}  
\usepackage{threeparttable}
\usepackage{caption}
\usepackage{tcolorbox}
\usepackage{authblk}
\usepackage{comment}           % comment blocks
\usepackage{natbib}
\usepackage{amsmath}

\title{Real-Time Sensing of Inaccessible Physical Fields via an Edge-Deployable Hardware-Portable Graph Neural Operator}
\author[1]{William Howes}
\author[1]{Jason Yoo}
\author[1]{Kazuma Kobayashi}
\author[4]{Subhankar Sarkar}
\author[1]{Farid Ahmed}
\author[1,3,4]{Souvik Chakraborty}
\author[1,2]{Syed Bahauddin Alam\textsuperscript{*}}

\affil[1]{Grainger College of Engineering, Nuclear, Plasma \& Radiological Engineering Department, University of Illinois Urbana-Champaign, Urbana, IL, USA
}

\affil[2]{National Center for Supercomputing Applications, Urbana, IL, USA
}

\affil[3]{Department of Applied Mechanics, Indian Institute of Technology Delhi, New Delhi, India
}

\affil[4]{Yardi School of Artificial Intelligence, Indian Institute of Technology Delhi}

\affil[*]{Corresponding author: \href{mailto:alams@illinois.edu}{alams@illinois.edu}}

\begin{document}
\maketitle

\begin{abstract}
Real-time inference of inaccessible interior physical fields from sparse boundary observations is a fundamental but unresolved problem in scientific machine learning, with direct relevance to safety-critical monitoring across many engineering applications. Existing neural operators achieve high accuracy but leave deployment to embedded edge platforms unaddressed. Here we introduce \textbf{VIRSO} (\textbf{V}irtual \textbf{I}rregular \textbf{R}eal-Time \textbf{S}parse \textbf{O}perator), the first neural operator with a unique spatial-spectral architecture that explicitly addresses edge-deployment hardware. VIRSO learns a nonlinear mapping from sparse, geometrically disjoint boundary inputs to spatially continuous interior multiphysics fields on irregular unstructured meshes through a spectral--spatial decomposition explicitly aligned with hardware execution: a compute-bound graph spectral pathway and a memory-bandwidth-bound spatial-aggregation pathway, each independently characterized on datacenter and embedded accelerators. The design reduces the inference energy--delay product by 29$\times$ relative to the vanilla graph-operator baseline (206\,J$\cdot$ms~$\to$~7.0\,J$\cdot$ms on an NVIDIA H200) and enables 17.0\,samples/s embedded inference on an NVIDIA Jetson Orin Nano within 7.06\,W board-level power, without modification. A mesh-density-adaptive graph construction strategy (V-KNN) simultaneously improves accuracy and reduces graph edge count by 34\%. Across three benchmarks with reconstruction ratios from 47:1 to 156:1, VIRSO achieves mean relative $L_2$ errors below 1\% with fewer parameters than operator baselines and delivers an inference speedup of $\approx 10^4$ times over the high-fidelity reference solver. To our knowledge, this is the first demonstration of a single-digit- watt neural operator, establishing hardware co-design as a missing ingredient in operator-based inference and a tractable path to real-time deployment.
\end{abstract}

% keywords can be removed

\section{Introduction}

%%%%%%%%%%% Paragraph-001: Universal inverse problem %%%%%%%%%%%

A defining limitation of contemporary sensing technology is that the physical quantities most critical to safety, performance, and decision-making are often precisely those that cannot be measured directly. In cardiovascular health monitoring, hemodynamic fields inside vessels and ventricles are inaccessible to continuous instrumentation while wall-pressure and surface acoustic measurements are accessible \cite{kobayashi2025proxies}. In subsurface resource assessment, formation flow and pressure fields cannot be directly observed while borehole-confined sensors provide sparse boundary data. In civil infrastructure monitoring, internal stress, fatigue, and corrosion fields evolve invisibly within bridges, pipelines, and pressure vessels while only surface-mounted strain gauges remain accessible \cite{hossain2024sensor}. In advanced industrial and nuclear systems, interior temperature, velocity, and turbulent transport fields develop throughout reaction zones that no physical sensor can survive. Across these domains, the structural form of the sensing problem is identical: continuous high-dimensional interior fields must be inferred from sparse boundary measurements, with reconstruction ratios that frequently exceed three to four orders of magnitude in underdetermination. No physical sensor can close this gap; no real-time physics-based simulation can either, because inverse reconstruction from sparse boundary data demands repeated high-fidelity solves at latencies incompatible with operational decision-making. The unmet need is therefore neither a better physical transducer nor a faster simulator, but a fundamentally new class of sensing instrument --- one whose function is to convert sparse boundary observations into complete interior field measurements in real time, within the power and latency constraints of deployed hardware.

\begin{tcolorbox}[colback=blue!10!white, colframe=blue!50!black, title=\textbf{Computational Sensing: A Deployable Sensing Modality}, coltitle=white, fonttitle=\bfseries] 
\vspace{-1mm}
We define \textbf{computational sensing} as the recovery of physically meaningful interior fields from sparse boundary measurements through learned operator mappings that satisfy deployment constraints --- latency, memory, and power --- thereby functioning as virtual sensing instruments. Computational sensing does not replace physical transducers; it extends them. Boundary sensors supply the observable signal; the computational sensing operator supplies the interior reconstruction that no physical transducer can deliver and no real-time simulator can produce. To function as a deployed sensing instrument rather than an offline surrogate, the operator must be co-designed with the compute-versus-memory-bandwidth hierarchy of the target hardware, making hardware constraints part of the sensing formulation, not post-processing.

\vspace{2mm}
\begin{center}
\renewcommand{\arraystretch}{1.15}
\begin{tabular}{lcc}
\toprule
\textbf{Sensing Capability} & \textbf{Conventional Sensing} & \textbf{Computational Sensing (VIRSO)} \\
\midrule
Direct measurement of interior fields & Required & Not required \\
Operation in inaccessible regions & Infeasible & Reconstructed from boundary data \\
Real-time field-level output ($<$1\,s) & N/A & Yes \\
Arbitrary irregular geometry & Limited & Yes \\
Embedded deployment ($<$10\,W) & Sensor-dependent & Yes (Jetson Orin Nano) \\
Retraining for new operating conditions & N/A & Not required \\
Reconstruction ratio (interior/boundary) & 1:1 & up to 156:1 \\
\bottomrule
\end{tabular}
\end{center}
\vspace{-1mm}
\end{tcolorbox}

%%%%%%%%%%% NEW: Nature Sensors scope alignment %%%%%%%%%%%

The framework developed here sits at the intersection of three active research directions in scientific machine learning and applied AI. First, as a learned-operator system, it extracts interior-field information from indirect boundary measurements without explicit governing-equation access, advancing the operator-learning literature into the regime of severely underdetermined inverse problems. Second, as a hardware-co-designed inference architecture, it imports the compute-versus-memory-bandwidth roofline analysis from compute-efficient deep learning and applies it to operator learning, where it has been absent. Third, as an embedded-deployment system, it demonstrates that physically meaningful PDE-governed inference can execute within single-digit-watt power envelopes on commodity edge hardware, opening operator-based inference to deployment regimes previously confined to compressed CNN or transformer workloads. The convergence of these three directions defines the contribution of this work and motivates its evaluation across both algorithmic-accuracy and hardware-deployment metrics.

%%%%%%%%%%% Paragraph-002: Domain instantiation %%%%%%%%%%%

This challenge is particularly acute in advanced nuclear energy systems, where it appears in its most demanding form. Small modular reactors (SMRs) and microreactors are attracting increasing attention as reliable and low-carbon energy sources for remote and high-demand industrial applications \cite{testoni2021review, kornecki2024role}. Safe and efficient operation requires continuous knowledge of internal reactor states, including temperature distributions, coolant velocity fields, and turbulent transport characteristics that govern both performance and safety margins. Yet the extreme environment inside an operating reactor, characterized by intense neutron flux, elevated temperatures, and restricted physical access, makes direct measurement of interior quantities largely infeasible. Operators must therefore infer the internal multiphysics state from a limited number of sensors located at accessible boundaries or external instrumentation points \cite{kobayashi2025proxies,kobayashi2025network,park2025bridging,kobayashi2024improved,hossain2024sensor,kobayashi2024deep}. The resulting reconstruction problem requires inferring a complete spatially distributed and physically coupled field from sparse boundary observations defined on highly irregular component-specific geometries.

\textbf{The sensing task addressed here is also categorically distinct from 
conventional virtual sensing and soft-sensing regression.} 
Methods commonly cited in that literature, including Support Vector Machines, 
Moving Horizon Estimators, Physics-Informed Neural Networks (PINNs), and 
sparse identification methods such as SINDy, are excluded from the 
quantitative comparison for the following precise reasons, not as an omission 
but as a principled methodological boundary:

\begin{itemize}
    \item \textbf{Regression-based virtual sensing} (SVMs, classical ML) \cite{DiNoia2018SVM, SCHILLER2026112790MHE}
    produces scalar or low-dimensional vector estimates at fixed, instrumented 
    locations. The output dimension is fixed at training time and cannot be 
    evaluated at arbitrary interior coordinates; the output space is 
    finite-dimensional by construction.
    
    \item \textbf{PINNs} \cite{RAISSI2019686PINN} embed the PDE residual as a training loss and must be 
    re-optimized from scratch for every new boundary condition instance. 
    This instance-specific retraining requirement is incompatible with the 
    real-time, train-once-deploy-continuously operational constraint central 
    to this work.
    
    \item \textbf{SINDy} \cite{Brunton_2016SINDY} identifies a parameterized differential equation from 
    trajectory data and requires explicit governing equations at inference time. 
    It provides no mechanism for field-level inference in domains where the 
    governing equations are unavailable or where the input and output domains 
    are geometrically disjoint ($X \cap Y = \emptyset$).
\end{itemize}

In addition to the fundamental virtual sensing problem addressed, advanced reactors (microreactors and SMRs) \cite{alam2019neutronic,alam2019small1,alam2019small2} intended for remote and off-grid siting impose a further constraint that the neural operator literature has not yet addressed directly: instrumentation and control systems on these platforms must operate within single-digit watt continuous power budgets on embedded accelerators, without access to datacenter-scale GPU infrastructure. Deployability under this hardware constraint is a necessary condition for practical virtual sensing, not an engineering convenience: a reconstruction operator that achieves sub-1\% field error but requires kilowatt-scale power draw cannot function as a deployed instrument.

%%%%%%%%%%% Paragraph-003: Digital twin limitations %%%%%%%%%%%

Digital twin (DT) frameworks \cite{daniell2025digital,kobayashi2024physics} have been proposed to address this monitoring challenge by constructing virtual replicas that evolve alongside the physical system \cite{daniell2025digital,iyengar14advances,HUANG2025126922, Hossain_Ahmed_Kobayashi_Koric_Abueidda_Alam_2025,kobayashi2024explainable,kobayashi2024ai}. In principle, a DT estimates unobservable interior quantities by repeatedly solving the governing multiphysics equations, typically coupled nonlinear partial differential equations discretized using high-fidelity numerical methods. In practice, however, each forward solve requires explicit governing equations and seconds to hours of computation, and inverse reconstruction from sparse boundary data demands either many repeated solves or large precomputed libraries that cannot adapt to evolving operating conditions. Real-time monitoring systems require field estimates on timescales of 
milliseconds to seconds without the need for complete understanding of governing equations, a regime that physics-based solvers cannot approach. 
Moreover, their computational structure is tightly coupled to iterative 
numerical schemes that do not map efficiently onto modern parallel hardware, 
making low-latency, energy-efficient deployment infeasible. A fundamentally 
different inference architecture is therefore required.

%%%%%%%%%%% Paragraph-004: Deep learning limitations %%%%%%%%%%%

Recent advances in machine learning offer a promising alternative. Deep neural networks have demonstrated remarkable capacity to approximate high-dimensional nonlinear mappings and have been widely used as surrogates for complex simulations in weather forecasting, structural analysis, and fluid dynamics \cite{Bi2023-tl, cai2021physicsinformedneuralnetworkspinns}. However, conventional neural network architectures learn mappings between fixed-resolution vectors and therefore remain tightly coupled to the spatial discretization of the training data \cite{jha2025theoryapplicationpracticalintroduction, JMLR:v24:21-1524}. Methods like Physics-Informed Neural Networks require re-optimization for new boundary conditions. Changing mesh resolution, sensor placement, or geometric configuration typically requires retraining from scratch. More fundamentally, these architectures are not designed for cross-domain reconstruction, where inputs are defined on a sparse boundary sensor set while outputs must be predicted on a dense interior mesh. This discretization dependence becomes a structural limitation for real-world monitoring systems. 

%%%%%%%%%%% Paragraph-005: Neural operators %%%%%%%%%%%

Neural operators have emerged as a principled framework for overcoming previous virtual sensin concerns. Rather than learning mappings between finite-dimensional vectors, neural operators approximate mappings between function spaces, enabling evaluation at arbitrary spatial coordinates and generalization across resolutions and geometries combined. As a result, the real-time requirements of virtual sensing is satisfied since no re-training is needed for unseen input conditions \cite{roy2026adversarial,Hossain_Ahmed_Kobayashi_Koric_Abueidda_Alam_2025, JMLR:v24:21-1524, jha2025theoryapplicationpracticalintroduction, Wang2025}. Some of the initially developed neural operator architectures include the Deep Operator Network (DeepONet) introduced a branch–trunk architecture that separates input encoding from spatial output decoding \cite{Lu2021-ci}. Spectral approaches such as the Fourier Neural Operator (FNO) learn operator kernels through global spectral convolutions on structured grids \cite{DBLP:journals/corr/abs-2010-08895}, while wavelet-based variants extend this concept using multiresolution representations \cite{tripura2022waveletneuraloperatorneural}. These frameworks have demonstrated strong performance as surrogate solvers for PDE systems. However, they remain primarily designed for forward operator settings in which inputs and outputs share the same domain structure and sensor coverage is dense relative to the output field. From a computational perspective, these neural operator architectures are dominated by dense linear transformations that map efficiently onto high-throughput GPU hardware. However, their extension to irregular domains requires more complex operations that bring with them fundamentally different computational characteristics. 

%%%%%%%%%%% Paragraph-006: Irregular geometry challenge %%%%%%%%%%%

Handling irregular geometries remains a major challenge for neural operators. Graph Neural Operators (GNOs) extend operator learning to unstructured meshes through message passing on graph representations \cite{DBLP:journals/corr/abs-2003-03485}. However, repeated neighborhood aggregation introduces over-smoothing that degrades high-frequency information in boundary layers and vortex structures. Spectral graph methods instead operate in the graph Fourier domain \cite{bruna2014spectralnetworkslocallyconnected, DBLP:journals/corr/DefferrardBV16}, but exact eigen-decomposition scales cubically with the number of nodes. Geometry-aware variants such as Geo-FNO attempt to embed irregular geometries into structured latent grids \cite{JMLR:v24:23-0064}, yet such diffeomorphic mappings are not guaranteed to exist for complex engineering geometries. Consequently, sparse-to-dense multiphysics reconstruction on irregular domains imposes three simultaneous constraints: geometric irregularity, sparse cross-domain input, and coupled multiphysics output.

%\textbf{Hardware--algorithm gap.} 
The challenge of sparse-to-dense reconstruction on irregular geometries is therefore not purely representational, but also computational. Architectures must simultaneously satisfy three constraints: (i) expressivity sufficient to resolve coupled multiphysics fields with highly irregular geometries, (ii) robustness to sparse and cross-domain sensing inputs, and (iii) computational structure compatible with low-latency execution on hardware platforms with limited memory bandwidth and power budgets. Existing graph-based methodologies can naturally address the first constraint in isolation but do not explicitly consider the interaction between operator structure and hardware execution, as well as sparse boundary input and full-field reconstruction. The consequences are quantifiable: the vanilla Graph Neural Operator requires 572\,W instantaneous GPU power and an energy-delay product of approximately 206\,J$\cdot$ms per inference sample on a modern datacenter accelerator: more than an order of magnitude above any embedded deployment budget and $20\times$ higher than the most efficient neural operator alternative on the same benchmark. Accuracy and hardware deployability cannot therefore be optimized independently; the reconstruction operator must be structured so that its computational pathway maps more efficiently onto resource-constrained hardware. Moreover, such operators must deal with sparse boundary input and high reconstruction ratios that are far above standard field to field neural operator benchmarks. To our knowledge, no prior neural operator, especially graph-based, explicitly addresses and analyzes both sparse-to-field reconstruction and hardware-constrained deployability while providing improved performance over existing operators within highly irregular geometries with complex multiphysics. 

None of the traditional virtual sensing/digital twin methods (Physics Solvers, SINDy, PINNs, etc.) can address the triple constraint that defines 
the realistic virtual sensing problem: geometric irregularity of the output domain, sparse 
cross-domain sensing with no spatial overlap between input and output, and simultaneous edge deployment. The appropriate baseline set is 
therefore drawn exclusively from the neural operator literature, specifically, 
the architectures most capable of irregular-geometry and multi-field 
reconstruction. Comparing against the above classical 
methods would require adapting them to a problem class they are not designed 
for and would not constitute a meaningful assessment of the capability gap 
this work addresses.

%%%%%%%%%%% Paragraph-007: VIRSO introduction %%%%%%%%%%%

\begin{tcolorbox}[colback=blue!10!white, colframe=blue!50!black, coltitle=white, fonttitle=\bfseries] Here we introduce the \textbf{Vi}rtual \textbf{I}rregular \textbf{R}eal-Time \textbf{S}parse \textbf{O}perator (\textbf{VIRSO}), a geometry-aware neural operator designed to reconstruct complete multiphysics field distributions from sparse boundary observations on unstructured domains. VIRSO builds on recent advances in graph-based neural operators, including the spatial–spectral graph neural operator (Sp$^2$GNO) \cite{SARKAR2025117659}, and extends them to sparse cross-domain reconstruction. A graph-based domain representation encodes irregular geometries without requiring structured grids. A latent input embedding projects sparse boundary observations into a shared representation that enables interior field inference. Finally, a spectral–spatial collaboration mechanism combines truncated graph Fourier convolutions with gated spatial aggregation to capture both long-range physical correlations and localized high-gradient features but without the high scalability of previous methodologies, enabling edge-feasibility for virtual sensing applications. Together these components define an operator mapping $\mathcal{G}: \mathbf{u}_{\mathrm{sparse}} \mapsto s(\mathbf{x})$ from sparse boundary observations to continuous interior multiphysics fields. \textbf{VIRSO introduces four key contributions:} 1. An irregular-geometry handling neural operator for sparse boundary-to-interior field reconstruction on irregular, cross-domain sensing configurations. 2. A spectral--spatial graph architecture whose dominant pathway (graph spectral convolution) is compute-bound and hardware-portable, while the auxiliary spatial branch provides local calibration. 3. A mesh-density-adaptive graph construction strategy (V-KNN) that improves reconstruction accuracy while reducing edge count by 34\% relative to the uniform high-connectivity alternative. 4. A validated edge-deployment pathway for real-time sensing that is hardware portable: all pretrained full VIRSO configurations execute on an NVIDIA Jetson Orin Nano at sub-second latency, with a sub-100 millisecond latency for the resource-constrained configuration, within 7-8\,W board-level power, without retraining, quantization, or architecture-specific modification. \end{tcolorbox}

%%%%%%%%%%% Paragraph-008: Variable-KNN %%%%%%%%%%%
Compared to prior neural operator frameworks (e.g., FNO, DeepONet, GNO, NOMAD, and recent variants), this work places a more explicit emphasis on hardware-aware design and edge deployability. While existing approaches do consider aspects of efficiency, scalability, or memory usage \cite{SARKAR2025117659, garg2023neuroscienceinspiredscientificmachine, garg2023neuroscienceinspiredscientificmachinepart1, Yin2024}, they are typically developed and evaluated with a primary focus on accuracy, generalization, and discretization invariance. In contrast, VIRSO not only incorporates deployment and scalability considerations into the architectural design but is also evaluated through inference on resource-constrained hardware, including an NVIDIA Jetson Nano, highlighting its suitability for edge settings. Beyond the operator architecture, we identify graph construction as an important and largely unexplored source of inductive bias in graph-based operator learning. Inspired by mesh refinement strategies in classical finite element analysis, we introduce \textbf{Variable-KNN (V-KNN)}, a mesh-density-adaptive graph construction strategy that assigns higher neighbor counts to nodes in regions of higher geometric complexity. This connectivity reorients the eigenmodes of the normalized graph Laplacian to align with underlying flow structures, improving reconstruction accuracy while maintaining edge efficiency.

%%%%%%%%%%% Paragraph-009: Demonstration %%%%%%%%%%%

We evaluate VIRSO on three multiphysics benchmarks of increasing geometric complexity drawn from nuclear engineering: a transient lid-driven cavity flow, a pressurized water reactor subchannel problem, and a wavy-insert heat exchanger \cite{AHMED2024104583} requiring reconstruction of coupled velocity and pressure fields. In the subchannel benchmark the reconstruction ratio reaches $51{:}1$, confirming that VIRSO addresses a genuinely underdetermined inverse problem rather than smooth spatial interpolation. Across all benchmarks VIRSO achieves mean relative $L_2$ errors below $1\%$ while using fewer parameters than competing neural operator baselines. These accuracy results are inseparable from hardware viability. VIRSO is, to our knowledge, the first neural operator whose architecture is co-designed with target-hardware execution constraints and validated end-to-end on embedded hardware. Hardware--algorithm co-design is a standard principle in compute-efficient deep learning for CNNs and transformers~\cite{10545889, ALSHARIF20251739, Kong2026-tu, Yao2025-mj, Meng2025}, but it has not been imported into operator learning prior to this work. The result is a quantified deployment regime that is structurally inaccessible to prior operator architectures: 17.0\,samples/s embedded inference at 7.06\,W board-level power on a Jetson Orin Nano, achieved without retraining or quantization. The full 10-layer VIRSO configuration achieves an energy-delay product of $1.30 \times 7.77 \approx 10.1$\,J$\cdot$ms on the H200, compared to $10.07 \times 20.48 \approx 206$\,J$\cdot$ms for the graph-based GNO baseline, and sustains 1.78\,samples/s at 7.58\,W board-level power on an NVIDIA Jetson Orin Nano without any model modification. The 10-layer VIRSO provides this computational performance along with highly accurate full-field reconstruction at real-time due to the efficient spectral-spatial design that allows for sophisticated graph analysis and computation requirements more favorable towards edge-constrained applications. The 2-layer lightweight configuration further achieves 1.95\% mean reconstruction error at 0.54\,J/it and 4.29\,ms latency, comparable to the energy and latency of full-scale Geo-FNO and NOMAD while degrading by $2$--$3\times$ in error under compression versus $4$--$8\times$ for those architectures. These results confirm that VIRSO is hardware-portable and satisfies the accuracy, energy, and latency constraints of real-time edge-deployed virtual sensing simultaneously, not as independent objectives. Moreover, the spectral-only VIRSO configuration, with its compute-bound architecture, sustains 17.0\,samples/s at 7.06\,W board-level power on the Jetson Nano with minimal degradation in performance. The success of the spectral model establishes an aspect of flexibility in VIRSO's architecture, unique among existing operator architectures, where extreme resource constrained applications can utilize the spectral-only analysis with low risk of reconstruction failure, especially if higher frequency modes are negligible. If more complex physics is present, the bandwidth-limited spatial layer might be required for accurate virtual sensing and calibration, resulting in $9.6\times$ more latency that is significantly lower than the vanilla Graph Neural Operator but still prompts further work towards hardware-based acceleration.

%%%%%%%%%%% Paragraph-010: Significance %%%%%%%%%%%
\textbf{The problem addressed in this work is categorically distinct from classical state estimation and dynamic observer design.} Kalman-filter families, Luenberger observers, and boundary PDE observers, such as those developed for traffic flow and ODE-hyperbolic PDE cascades, are dynamic systems that estimate a finite-dimensional state vector evolving in time under a known, explicitly specified governing equation, with time as the free variable and model structure as a prerequisite \cite{FERRANTE2020109027EX1, 10192343EX2, 9086794EX3}. The problem we address is structurally different in every relevant dimension: the target quantities are spatially continuous field distributions defined on two-dimensional irregular domains with up to 3,977 nodes; the input observations are defined on a geometrically disjoint domain with no spatial overlap with the output ($X \cap Y = \emptyset$); no closed-form governing equation is available at inference time; and evaluation must occur at arbitrary interior coordinates that are physically inaccessible to instrumentation. This is the problem of learning a nonlinear operator $G : U \to L^2(Y; \mathbb{R}^k)$ between infinite-dimensional function spaces, a mathematical object that classical state observers are not designed to estimate and do not claim to estimate. Positioning the contribution within operator learning rather than state estimation is therefore not a framing choice but a mathematical necessity: the output of VIRSO is an element of a function space, not a finite-dimensional vector, and no reformulation of Kalman filtering, MHE, or PDE boundary observer design provides a mechanism to infer arbitrary-resolution interior fields on irregular unstructured meshes without an explicit forward model and without retraining for each new boundary condition instance.

%%%%%%%%%%% Paragraph-010b: Preemptive clarification for R3-type concerns %%%%%%%%%%%

In summary, VIRSO extends graph-based neural operator learning to the general problem of multiphysics state inference from sparse cross-domain observations on irregular geometries, while explicitly addressing the computational constraints of real-world deployment. By eliminating the need for repeated high-fidelity simulations during inference and enabling hardware-efficient execution, the framework supports real-time reconstruction of complex physical fields within practical latency and energy budgets. Although demonstrated for nuclear thermal–hydraulic systems, the same triple constraint arises in many domains \cite{gupta2025continuous}, including cardiovascular flow monitoring, subsurface resource assessment, structural health diagnostics, and atmospheric sensing. In this work, we introduce VIRSO, a neural operator architecture whose spatial--spectral decomposition is explicitly designed for hardware co-design, enabling real-time virtual sensing on embedded edge hardware. More broadly, VIRSO is a step toward a deployment-aware paradigm for scientific machine learning, in which the operator and the hardware on which it executes are designed jointly, and benchmark accuracy is evaluated alongside latency, energy, and embedded feasibility as primary, not secondary, metrics.

\section{Mathematical \& Algorithm Formulation}

The fundamental sensing challenge is to recover a continuous physical field $\mathbf{s}(\mathbf{x})$ throughout an unobservable interior domain $\mathcal{Y}$ from a finite set of measurements $\{u_i\}_{i=1}^M$ confined to an accessible boundary $\mathcal{X}$, with $\mathcal{X} \cap \mathcal{Y} = \emptyset$. This is the canonical virtual sensor problem: inferring what cannot be directly measured from what can. Classical sensor fusion approaches assume either a known forward model or dense spatial coverage; VIRSO eliminates both assumptions by learning the boundary-to-interior operator directly from data, essentially mapping scalar and functional boundary inputs to multiple spatially distributed physical outputs. 

Let $\mathcal{Y} \subset \mathbb{R}^d$ represent our system's geometrical $d$-dimensional spatial domain that we wish to predict the functional output on. VIRSO attempts to solve the governing PDEs represented through the following operator formulation:

\begin{equation}
\mathcal{G}: \mathcal{U} \to \mathcal{S}, \quad \mathcal{G}(\mathbf{u})(\mathbf{x}) = \mathbf{s}(\mathbf{x}), \quad \mathbf{x}\in\mathcal{Y},
\end{equation}

where $\mathcal{G}$ represents our nonlinear operator between functional spaces, $\mathbf{u} = [u_1,...,u_b] \in \mathcal{U}$ depicting $b$ multi-modal input spaces, where $\mathcal{U}=\prod_{i=1}^b \mathcal{F}_i$, and $u_i$ can be a scalar $\mathbb{R}$ or a functional space $L^2(D')$ defined on a separate domain. The output $\mathbf{s(x)}\in\mathbb{R}^k$ defines $k$ physical quantities on the desired domain $\mathcal{Y}$, such that the output space is $\mathcal{S}\subset L^2(\mathcal{Y};\mathbb{R}^k)$.

Kernel-based neural operators approximate $\mathcal{G}$ with an iterative convolution integral formation with nonlinear activation described by the following Green's function inspired framework \cite{JMLR:v24:21-1524, Zappala2024}:

\begin{equation} \label{eq:kernel_int}
\mathbf{v}_{t+1}(\mathbf{x}) = \sigma(W\mathbf{v}_t(\mathbf{x}) + \int_{\mathcal{Y}}{\mathcal{K}_\phi}(\mathbf{x},\mathbf{z})\mathbf{v}_t(\mathbf{z})d\mathbf{z}),
\end{equation}

where $\mathbf{v}_t$ and $\mathbf{v}_{t+1}$ represent intermediate function evaluations on the desired domain $\mathbf{x} \in \mathcal{Y}$ with $t$ ranging from $0$ to $T$ iterative layers, $\sigma$ representing a nonlinear activation such as Sigmoid or ReLU, $W \mathbf{v}_t(\mathbf{x})$ represents a weighted residual connection, and $\mathcal{K}_\phi$ represents the parameterized kernel utilized in the integral over the geometric domain $D$. FNO, WNO, and graph-based methods attempt to estimate the kernel integral utilizing the domain discretization either through spectral convolutions which project to the spectral domain, simplifying the integral to matrix multiplication, or local spatial convolutions which directly estimate the integral through summation.

\begin{figure}[htbp]
    \centering
    \includegraphics[width=0.95\textwidth]{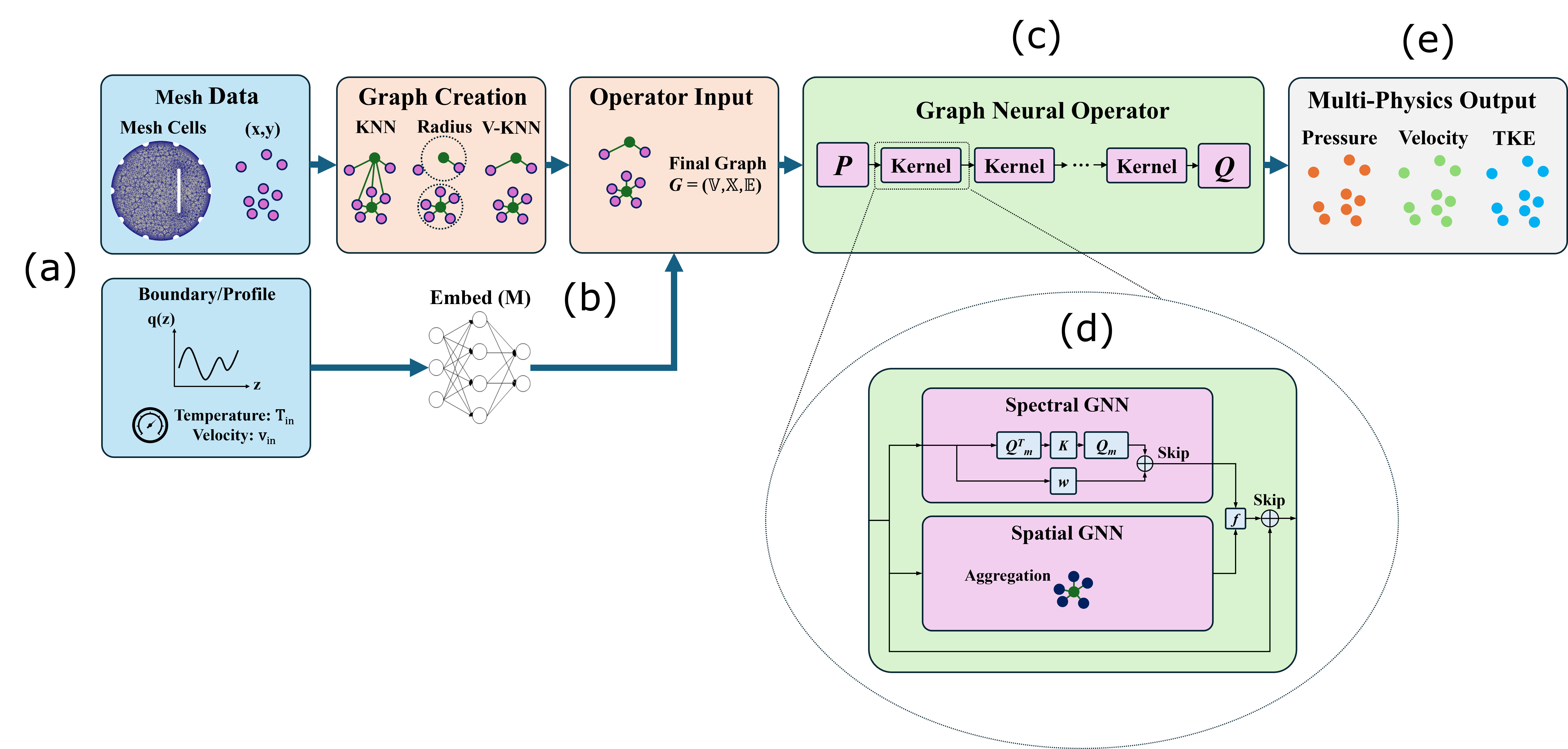}
    \caption{
    \textbf{Multi-Physics Virtual Sensing Framework for VIRSO}. 
    \textbf{(a)} Training and validation for the VIRSO model originates from mesh-based numerical solvers. The boundary conditions at the geometry inlet and a discretization of the forcing profile (or signal for transient) are given as initial input to our framework along with the generated mesh which is converted to positional coordinates of each mesh cell throughout the domain of the desired multi-physics output.
    \textbf{(b)} The mesh cell coordinates become our graph nodes $\mathbb{V}$ and utilizing graph creation methods such as K-Nearest Neighbors (KNN), Radius, or a custom Variable KNN (V-KNN), we can define crucial edge connections $\mathbb{E}$ between node points. The input signal and values are mapped to a latent embedding space and then copied to each node and combined with node coordinates to define the initial node features $\mathbb{X}$.
    \textbf{(c)} The VIRSO model projects the input into a hidden function dimension and repeatedly extracts global and local features through spectral-spatial blocks. Eventually the operator downlifts to the final output space.
    \textbf{(d)} A close up into the VIRSO block architecture. The GFT allows for the spectral convolution of the first $m$ modes, and the spatial aggregation GNN provides local feature extraction. The collaboration projection $f$ prevents over-smoothing while skip connections assist training and also allow local information flow.
    \textbf{(e)} Defined on the mesh-node grid, we have multiple predicted field outputs that describe physics within our domain such as pressure, velocity, and turbulent kinetic energy (TKE).
    }
    \label{fig:overall_framework}
\end{figure}

Figure \ref{fig:overall_framework} summarizes how VIRSO handles the defined problem statement and approximates the kernel convolution integral for operator learning. Essentially, VIRSO embeds boundary information with an embedding layer $M$ which can be a Fully Connected Network (FCN) \cite{SCABINI2023128585} or other network model. This latent information is combed with coordinates to form the node features $\mathbb{X}$ for the KNN/Radius/V-KNN constructed graph ($\mathbb{V}, \mathbb{E}$) and fed through a projection mapping $P$ and $T$ sequential kernel layers mirroring Eq. \ref{eq:kernel_int}. Each layer performs a global spectral analysis and local spatial analysis which both attempt to approximate the kernel integral (only the spectral layer contains a weighted residual and activation function). Information from both approximations is combined through a projection mapping $f$ (linear or non-linear FCN) with an additional residual skip for the next layer, allowing a full-scale analysis without over-smoothing and scalability concerns. Lastly, a downlift layer $Q$ projects the intermediate features to the final multiphysics output. Further information regarding the algorithm details of VIRSO is located in the Methods section \ref{sec:VIRSO_algo}.

% ============================================================
\section{Results}
% ============================================================
\textbf{We evaluate VIRSO not as a predictive model but as a deployable sensing instrument, characterizing it on three axes that jointly define its operational suitability: sensing accuracy at high reconstruction ratios, latency under embedded hardware constraints, and board-level power within single-digit-watt envelopes.} The baseline set is drawn from the neural operator literature because no other class of methods reconstructs spatially continuous field distributions from sparse cross-domain boundary observations on irregular geometries. Classical regression-based virtual sensing approaches, including Support Vector Machines, Kalman filters, and Moving Horizon Estimators, produce scalar or low-dimensional vector estimates at instrumented locations; they cannot reconstruct field distributions at arbitrary interior coordinates in domains with no instrumentation, because their output space is finite-dimensional and fixed at training time. Physics-Informed Neural Networks (PINNs) embed the PDE residual as a training objective and must be re-optimized from scratch for every new boundary condition instance, incurring training-time costs that are incompatible with real-time sensing and precluding deployment under the operational constraints considered here. SINDy identifies parameterized differential equations from trajectory data and provides no mechanism for field inference in domains where the governing equations are unknown at inference time. None of these methods can address the triple constraint that defines our problem: geometric irregularity of the output domain, sparse cross-domain sensing inputs with no spatial overlap with the output, and reconstruction ratios between 47:1 and 156:1. The appropriate baseline set is therefore drawn from the neural operator literature, specifically the architectures most capable of handling irregular geometries and multi-field outputs, which is precisely the set evaluated: GeoFNO, NOMAD, and GNO. Including the above classical methods as baselines would require adapting them to a problem class they are not designed for; doing so would not constitute a fair comparison and would not inform the question of which operator architecture is most capable for this sensing task. We evaluate VIRSO not only as a predictive operator but as a deployable sensing mechanism, focusing on accuracy, computational efficiency, and hardware realizability.

\begin{tcolorbox}[colback=gray!8!white, colframe=blue!50!black, title=\textbf{System-Level Significance: From Algorithm to Deployed Inference Instrument}, coltitle=white, fonttitle=\bfseries]
\vspace{-1mm}
The contribution of this work spans the algorithm--hardware--system stack. At the algorithmic level, VIRSO introduces a spectral--spatial graph operator co-designed with the compute-versus-memory-bandwidth hierarchy of deployment hardware. At the hardware level, the same pretrained model executes on a 90\,W datacenter GPU (H200) and a single-digit-watt embedded accelerator (Jetson Orin Nano) without retraining, quantization, or architectural change. At the system level, this enables a deployed inference instrument with sub-second latency, sub-10\,W power, and sub-1\% mean reconstruction error on irregular geometries with up to 156:1 underdetermination. The novelty is not algorithmic in isolation: it is the demonstration that operator learning can be designed with hardware realizability as a first-class constraint and validated end-to-end on embedded hardware, a regime that prior operator-learning work has not reached. The quantified system-level outcomes are summarized below.

\vspace{2mm}
\begin{center}
\renewcommand{\arraystretch}{1.15}
\begin{tabular}{lc}
\toprule
\textbf{System-Level Outcome} & \textbf{Value} \\
\midrule
Energy--delay-product reduction vs.\ graph-operator baseline (H200) & $29\times$ \\
Embedded throughput (Jetson Orin Nano, spectral-only) & 17.0\,samples/s \\
Embedded board-level power (VDD\_IN rail) & 7.06\,W \\
Inference speedup vs.\ ANSYS Fluent reference solver (LDC) & $>10{,}000\times$ \\
Reconstruction ratio range (interior values per boundary reading) & 47:1 -- 156:1 \\
Mean relative $L_2$ error across all three benchmarks & $<$\,1\% \\
Graph edge-count reduction via V-KNN at matched accuracy & 34\% \\
Retraining required for cross-platform deployment & None \\
Quantization or pruning required for embedded inference & None \\
\bottomrule
\end{tabular}
\end{center}
\vspace{-1mm}
\end{tcolorbox}

\subsection{The Computational Sensing Problem and Sensor-System Configuration}

VIRSO is a sensor-system component, not a surrogate solver. In the deployed configuration, accessible-boundary sensors --- inlet thermocouples, flow probes, and pressure transducers --- provide sparse boundary observations, and VIRSO converts those boundary streams into complete interior multiphysics field estimates in real time. We evaluate this configuration on three benchmarks of increasing sensing difficulty: the Lid-Driven Cavity (LDC) provides a controlled baseline on a uniform grid; the PWR Subchannel introduces an irregular geometry with cross-domain (axial-to-transverse) sensing; and the wavy-insert Heat Exchanger represents the most demanding configuration, combining maximum geometric irregularity, four coupled output channels, and a 156:1 reconstruction ratio --- a regime in which 156 interior field values must be recovered for every boundary sensor reading. The task is therefore a genuine sensing problem, not a surrogate acceleration problem. In the forward computation, a high-fidelity solver accepts a complete specification of boundary conditions and geometry as input and produces the spatially resolved interior field by discretizing and solving the governing partial differential equations. The forward evaluation based on traditional physics solvers for these use cases is accurate but computationally prohibitive for real-time use, requiring 36 minutes per sample on modern computing hardware for the Lid-Driven Cavity. VIRSO addresses the inverse: given only sparse observations from boundary sensors, a small number of scalar inlet values and a discretized forcing profile, which recover the complete interior multiphysics field consistent with those observations. This is the canonical virtual sensor architecture: the instrument reports what is accessible at the boundary; the model infers what is not accessible in the interior.

The difficulty is fundamental to the sensing configuration. For a given set of sparse boundary readings, many distinct interior field configurations are mathematically consistent with the same sensor data. Selecting the physically correct one requires the model to have internalized the governing physics through training. This is distinct from spatial interpolation, which assumes a known functional form connecting measurement locations, and from field-to-field regression, which assumes dense concurrent input and output measurements. The information asymmetry is extreme: in the geometry-specific reconstruction ratios evaluated below, between $47{:}1$ and $156{:}1$ interior field values must be recovered per boundary sensor reading.

We formalize that the operator VIRSO learns as follows. Let $\mathcal{Y} \subset \mathbb{R}^d$ denote the irregular interior domain discretized at $N$ spatial nodes, and let $\mathbf{u}_{\mathrm{bc}} \in \mathbb{R}^{M}$ denote the sparse multimodal boundary observations with $M \ll N$. The target output is $\mathbf{s} \in \mathbb{R}^{N \times C}$, where $C$ denotes the number of coupled physical quantities. VIRSO learns
\begin{equation}
\mathcal{G}: \mathbf{u}_{\mathrm{bc}} \mapsto \mathbf{s}(\mathbf{x}), \quad \mathbf{x} \in \mathcal{Y},
\end{equation}
directly from simulation data, without access to the governing equations at inference time and without any spatial overlap between the sensor domain and the output domain ($\mathcal{X} \cap \mathcal{Y} = \emptyset$ for all benchmarks). The reconstruction ratio $N \times C / M$ defines the sensing underdetermination directly. Across the three benchmarks evaluated here, this ratio ranges from $47{:}1$ (LDC) to $51{:}1$ (PWR Subchannel) to $156{:}1$ (Heat Exchanger), spanning the regime of genuine inverse sensing problems rather than smooth interpolation tasks.

VIRSO was evaluated against three neural operator baselines: the geometry-aware Fourier Neural Operator (Geo-FNO)~\cite{JMLR:v24:23-0064}, the Nonlinear Manifold Decoder for Operator Learning (NOMAD)~\cite{seidman2022nomadnonlinearmanifolddecoders}, and a vanilla Graph Neural Operator (GNO)~\cite{DBLP:journals/corr/abs-2003-03485}. These represent the strongest available candidate architectures for irregular-domain operator learning. All models were evaluated on an identical 80/20 train-test split using mean relative $L_2$ error as the primary metric. Full training hyperparameters are provided in Methods~(\ref{sec:training}). Uniquely among neural operator evaluations, the benchmarks are also analyzed under hardware deployment constraints: the spectral–spatial decomposition is analyzed in the context of compute-versus-memory-bandwidth hierarchy of both datacenter and embedded accelerators along with its flexibility in configuration and implementation (e.g. spectral-only), establishing that graph-based operator architectures can be designed with hardware execution characteristics in mind, in addition to predictive accuracy, and emphasizing how the VIRSO architecture moves towards edge-deployability for virtual sensing.

%%% SUBSECTION 1 %%%
\subsection{Lid-Driven Cavity: Transient Multiphysics 
Operator Learning on a Regular Domain}

The Lid-Driven Cavity (LDC) problem serves as an initial 
evaluation of VIRSO's ability to reconstruct coupled 
multiphysics fields from a time-varying boundary input. 
Although defined on a uniform grid, the LDC represents 
a genuinely hard operator learning problem: the input is 
a one-dimensional forcing signal $V(t)$ evaluated over 
90 discrete time steps, while the target output consists 
of three coupled fields — pressure $p(x,y)$, velocity 
magnitude $\|\boldsymbol{v}(x,y)\|$, and turbulent kinetic 
energy $k(x,y)$ — evaluated over 4,225 interior nodes. 
This corresponds to a reconstruction ratio of approximately 
$47{:}1$, with the three-channel output increasing the 
effective underdetermination further. The governing 
physics is described by the incompressible 
Reynolds-Averaged Navier-Stokes (RANS) equations 
with a $k$-$\varepsilon$ turbulence closure:

\begin{equation}
\nabla \cdot \boldsymbol{v} = 0,
\end{equation}
\begin{equation}
\frac{\partial \boldsymbol{v}}{\partial t} + 
(\boldsymbol{v} \cdot \nabla)\boldsymbol{v} = 
-\nabla p + \nabla \cdot \left[ \left( \nu + \nu_t \right) 
\left( \nabla \boldsymbol{v} + \nabla \boldsymbol{v}^\top 
\right) \right],
\end{equation}
\begin{equation}
\frac{\partial k}{\partial t} + \boldsymbol{v} \cdot 
\nabla k = P_k - \varepsilon + \nabla \cdot 
\left[ \left( \nu + \frac{\nu_t}{\sigma_k} \right) 
\nabla k \right].
\end{equation}

\begin{figure}[htbp]
    \centering
    \includegraphics[width=0.95\textwidth]{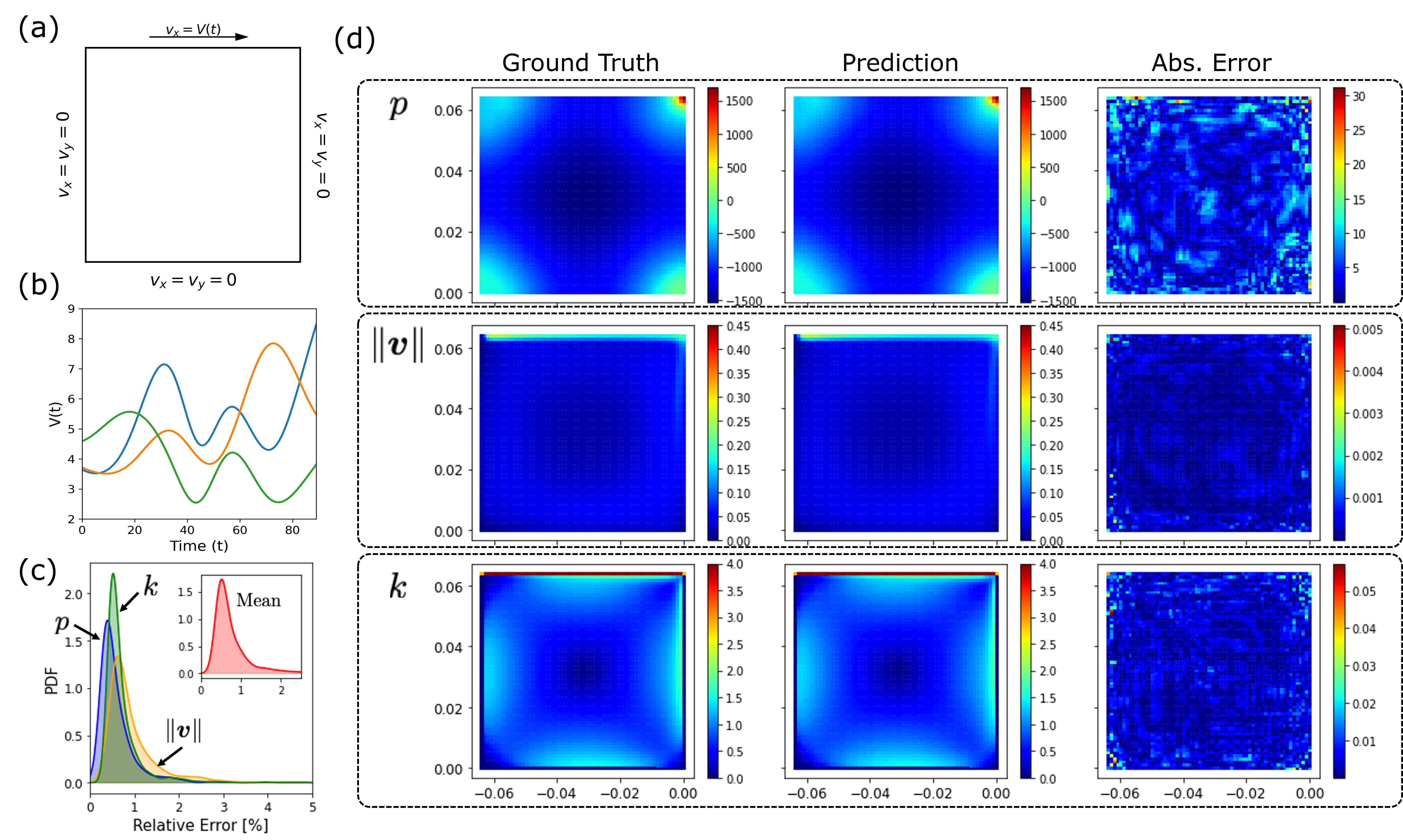}
    \caption{
    \textbf{Lid-Driven Cavity: Problem Formulation and Results.} 
    \textbf{(a)} Boundary conditions: no-slip at all walls except 
    the driven lid at $y=1$, where a time-varying forcing 
    signal $V(t)$ prescribes the horizontal velocity. 
    \textbf{(b)} Representative forcing signals $V(t)$ 
    evaluated at 90 discrete time steps, spanning a broad 
    range of transient dynamics.
    \textbf{(c)} Percentile error distributions from VIRSO 
    for pressure, velocity magnitude, TKE, and overall mean 
    error (blue, orange, green, red). All distributions are 
    centered below $1\%$ with low spread, confirming consistent 
    generalization across the test distribution.
    \textbf{(d)} 50th-percentile reconstruction: predicted 
    fields, ground truth, and absolute error for pressure, 
    velocity magnitude, and TKE. VIRSO accurately recovers 
    the primary vortex, shear layer, and corner eddies. 
    Elevated absolute error near boundaries and corners 
    reflects the sensitivity of turbulent physics to 
    regions underresolved by the uniform output grid.
    }
    \label{fig:ldc}
\end{figure}

The operator learned by VIRSO for this problem is:
\begin{equation}
\mathcal{G}_{ldc}: L^2(\mathbb{R}_{\geq0}) \to 
(L^2(\mathcal{Y}_{ldc}))^3, \quad 
\mathcal{Y}_{ldc} \subset \mathbb{R}^2,
\end{equation}
approximated as the discrete mapping 
$\mathcal{G}_{ldc}: \mathbb{R}^{90} \to \mathbb{R}^{4225\times3}$. 
The boundary conditions, domain geometry, and operator 
inputs are illustrated in Figure~\ref{fig:ldc}a,b. Details regarding model architecture choices are found in Methods \ref{sec:training}.

\textbf{Performance.} As shown in Table~\ref{tab:ldc_model_performance}, a 7 integral-kernel layer VIRSO achieves a mean relative $L_2$ error of $0.76\%$ across 
all three output fields, outperforming NOMAD ($0.86\%$), 
Geo-FNO ($1.11\%$), and GNO ($11.98\%$) at comparable or 
lower parameter count. The error distributions, shown in 
Table \ref{tab:ldc_model_percentile} of the Supplementary Material and Figure~\ref{fig:ldc}c, 
are narrow and unimodal across all fields, with interquartile 
ranges well below $1\%$. Turbulent kinetic energy exhibits 
the largest variance among the three outputs, consistent with 
its sensitivity to boundary-layer dynamics near the domain 
corners where the uniform grid provides insufficient resolution. 
The 50th percentile reconstruction (Figure~\ref{fig:ldc}d) 
accurately captures the primary recirculating vortex, the 
shear layer beneath the driven lid, and the secondary corner 
eddies. Absolute error is highest near the boundary and 
corner regions, consistent with the known sensitivity 
of turbulent flow to local geometric curvature.

The uniform $65 \times 65$ grid underresolves the turbulent corner eddies that are physically significant and are the primary discriminator between operator architectures. VIRSO's advantage is expected to increase with proper adaptive mesh refinement near boundary layers, a condition that is satisfied by the subsequent, more realistic benchmarks. With improved architectural choices — a GeLU activation after the spatial block, reordered normalization layers, and a weighted graph Laplacian — the mean error decreases to $0.65\%$, widening the margin beyond all benchmarks ($> 0.86\%$) without increasing parameter count or training time.

\begin{table}[htbp]
\centering
\caption{Model Comparison: Lid-Driven Cavity}
\begin{tabular}{@{}cccccc@{}}
\toprule
\multirow{2}{*}{\textbf{Model}} & 
\multirow{2}{*}{\textbf{Parameters}} & 
\multicolumn{4}{c}{\textbf{Mean Relative $L_2$ Error (\%)}} \\ 
\cmidrule(l){3-6} 
& & $p(x,y)$ & $\|\textbf{v}(x,y)\|$ & $k(x,y)$ & Mean \\ 
\midrule
\textbf{NOMAD}   & 1.63 M & 0.65 & 1.18 & 0.74 & 0.86 \\
\textbf{Geo-FNO} & 1.68 M & 0.94 & 1.08 & 1.32 & 1.11 \\
\textbf{GNO}     & 0.23 M & 2.84 & 10.89 & 22.23 & 11.98 \\
\textbf{VIRSO}   & 1.62 M & \textbf{0.62} & \textbf{0.95} & 
\textbf{0.71} & \textbf{0.76} \\ 
\bottomrule
\end{tabular}
\label{tab:ldc_model_performance}
\end{table}

%%% SUBSECTION 2 %%%
\subsection{PWR Subchannel: Multiphysics Inverse Reconstruction 
on an Irregular Geometry}

The PWR subchannel problem constitutes the first test of 
VIRSO under the full triple constraint. The output domain 
$\mathcal{Y}_{sub}$ is an irregular two-dimensional axial 
cross-section of a reactor subchannel, discretized at 
1,733 nodes by the high-fidelity ANSYS Fluent solver. 
The input consists of two scalar boundary conditions 
— inlet temperature $T_{in} \in \mathbb{R}$ and axial 
velocity $v_{in} \in \mathbb{R}$ — together with a 
one-dimensional axial heat source profile 
$q(z) = A\sin(\pi z / H)$ evaluated at 100 discrete 
axial positions along the entire three-dimensional geometry, for a total of 102 input values. 
The reconstruction ratio for this problem is:
\begin{equation}
\frac{N \times C}{M} = \frac{1{,}733 \times 3}{102} 
\approx 51{:}1,
\end{equation}
where $C=3$ output fields — velocity magnitude 
$\|\boldsymbol{v}(x,y)\|$, temperature $T(x,y)$, 
and turbulent kinetic energy $k(x,y)$ — are coupled 
through the governing RANS and energy equations. 
The input observations are defined on the axial 
direction ($z$-axis), while the output fields are 
defined on the transverse cross-section ($x$-$y$ plane) 
at a single axial point $z = z_0$: the input and output domains are 
geometrically disjoint. This cross-domain structure 
makes standard operator architectures architecturally 
misaligned with the problem, as their formulations 
presuppose coincident input and output domains.

The operator learned is:
\begin{equation}
\mathcal{G}_{sub}: (\mathbb{R})^2 \times L^2(\mathbb{R}) 
\to (L^2(\mathcal{Y}_{sub}))^3, \quad 
\mathcal{Y}_{sub} \subset \mathbb{R}^2,
\end{equation}
approximated as 
$\mathcal{G}_{sub}: \mathbb{R}^{102} \to \mathbb{R}^{1733\times3}$. Details regarding model architecture choices are found in Methods \ref{sec:training}.

\textbf{Performance.} Table~\ref{tab:subchannel_model_performance} 
shows that a 4 integral-kernel layer VIRSO achieves a mean error of $0.51\%$ — 
the best among all models, with a parameter count of 
only 0.34 M compared to NOMAD's 0.42 M and 
Geo-FNO's 2.70 M. Crucially, the performance advantage 
is largest for turbulent kinetic energy ($0.88\%$ vs. 
NOMAD's $0.96\%$), the output field most sensitive 
to local geometric complexity near the reactor rod surfaces. 
This pattern is consistent with VIRSO's design: 
the spectral-spatial collaboration mechanism is 
specifically intended to preserve local high-frequency 
information near geometrically complex boundaries, 
which TKE requires more than the smoother pressure 
and temperature fields.

\begin{figure}[htbp]
    \centering
    \includegraphics[width=0.95\textwidth]{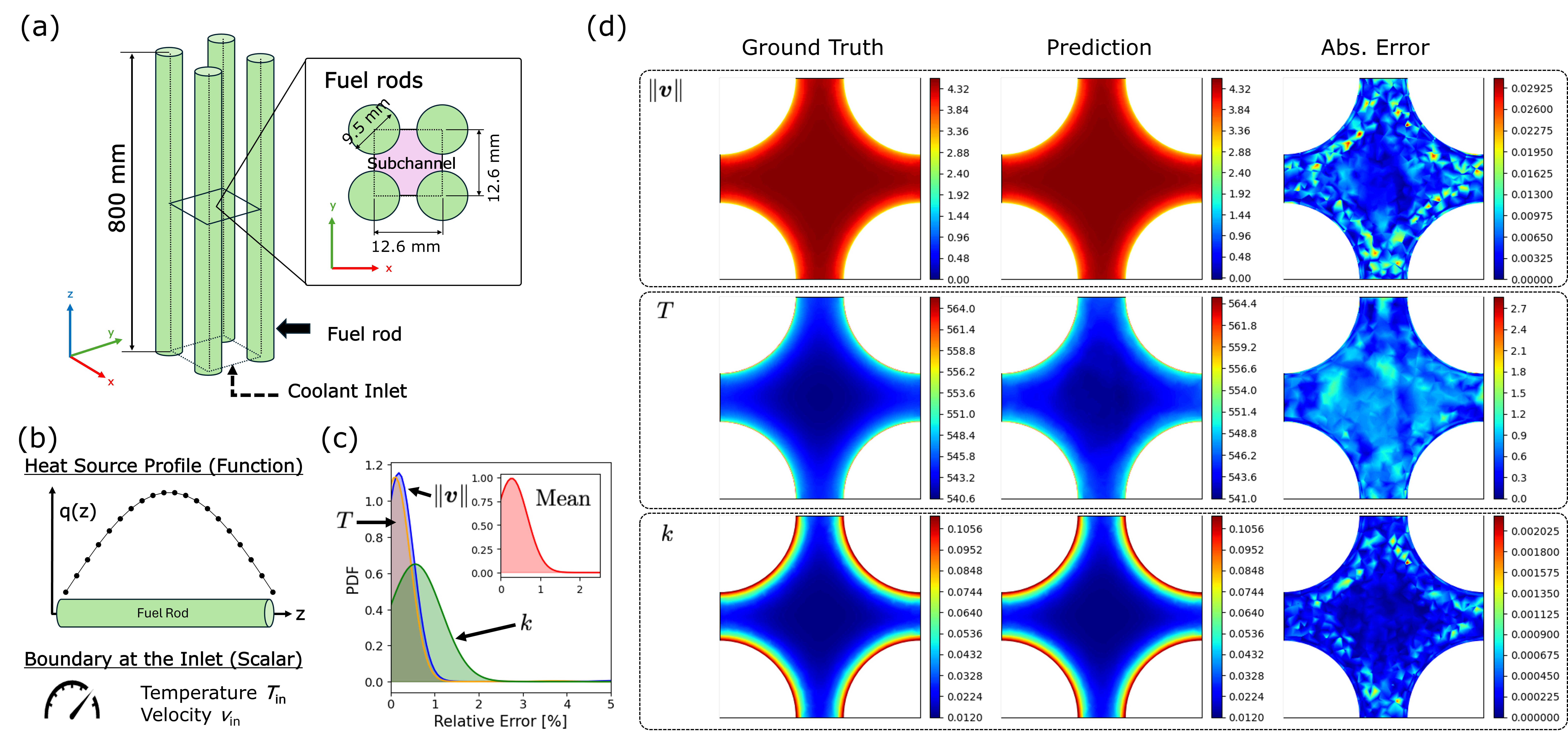}
    \caption{
    \textbf{PWR Subchannel: Problem Formulation and Results.} 
    \textbf{(a)} Geometry: coolant flow between four reactor 
    rods, analyzed at a fixed axial cross-section. No-slip 
    boundary conditions are enforced on rod surfaces; 
    inlet is at $z=0$.
    \textbf{(b)} Multimodal boundary inputs: axial heat 
    source profile $q(z)$ at 100 positions and two scalar 
    inlet conditions. These inputs are defined on the 
    axial direction, while outputs are defined on the 
    transverse cross-section — a geometrically disjoint 
    cross-domain configuration.
    \textbf{(c)} Percentile error distributions 
    (orange: temperature, blue: velocity, green: TKE, 
    red: mean). All distributions centered below $1\%$ 
    with narrow spread. 
    \textbf{(d)} 50th-percentile reconstruction. 
    VIRSO accurately captures wall shear layers, 
    central velocity structure, and symmetric 
    thermal gradient. Weak spatial correlation 
    of error with boundary proximity for velocity 
    and TKE; stochastic fluctuations for temperature.
    }
    \label{fig:subchannel}
\end{figure}

The percentile statistics in 
Table \ref{tab:subchannel_model_percentile} of the Supplementary Material demonstrate 
that VIRSO achieves consistently superior performance 
from the best-case to the 95th-percentile sample, 
with the narrowest interquartile range of all models. 
All distributions are centered below $1\%$, and the 
50th-percentile reconstruction (Figure~\ref{fig:subchannel}d) 
accurately captures wall shear layers near the reactor 
rod surfaces, the central velocity structure, and the 
symmetric thermal gradient. Absolute error shows weak 
spatial correlation with boundary proximity for velocity 
and TKE, and no systematic pattern for temperature, 
indicating predominantly stochastic model fluctuations 
rather than structural reconstruction failures. 
With further architectural refinements (nonlinear 
collaboration function, GeLU activation, weighted 
Laplacian), the mean error decreases to $0.43\%$ 
at 0.36 M parameters, further extending the advantage 
over all baselines.

\begin{table}[]
\caption{Model Comparison: PWR Subchannel}
\centering
\begin{tabular}{@{}cccccc@{}}
\toprule
\multirow{2}{*}{\textbf{Model}} & 
\multirow{2}{*}{\textbf{Parameters}} & 
\multicolumn{4}{c}{\textbf{Mean Relative $L_2$ Error (\%)}} \\ 
\cmidrule(l){3-6} 
& & $\|\boldsymbol{v}(x,y)\|$ & $T(x,y)$ & $k(x,y)$ & Mean \\ 
\midrule
\textbf{NOMAD}   & 0.42 M & 0.39 & \textbf{0.27} & 0.96 & 0.54 \\
\textbf{Geo-FNO} & 2.70 M & 1.55 & 1.06 & 2.88 & 1.83 \\
\textbf{GNO}     & 0.27 M & 4.99 & 0.50 & 3.47 & 2.99 \\
\textbf{VIRSO}   & 0.34 M & \textbf{0.37} & 0.28 & 
\textbf{0.88} & \textbf{0.51} \\ 
\bottomrule
\end{tabular}
\label{tab:subchannel_model_performance}
\end{table}

%%% SUBSECTION 3 %%%
\subsection{Wavy-Insert Heat Exchanger: Four-Field 
Reconstruction on a Highly Irregular Domain}

\begin{figure}[htbp]
    \centering
    \includegraphics[width=0.95\textwidth]{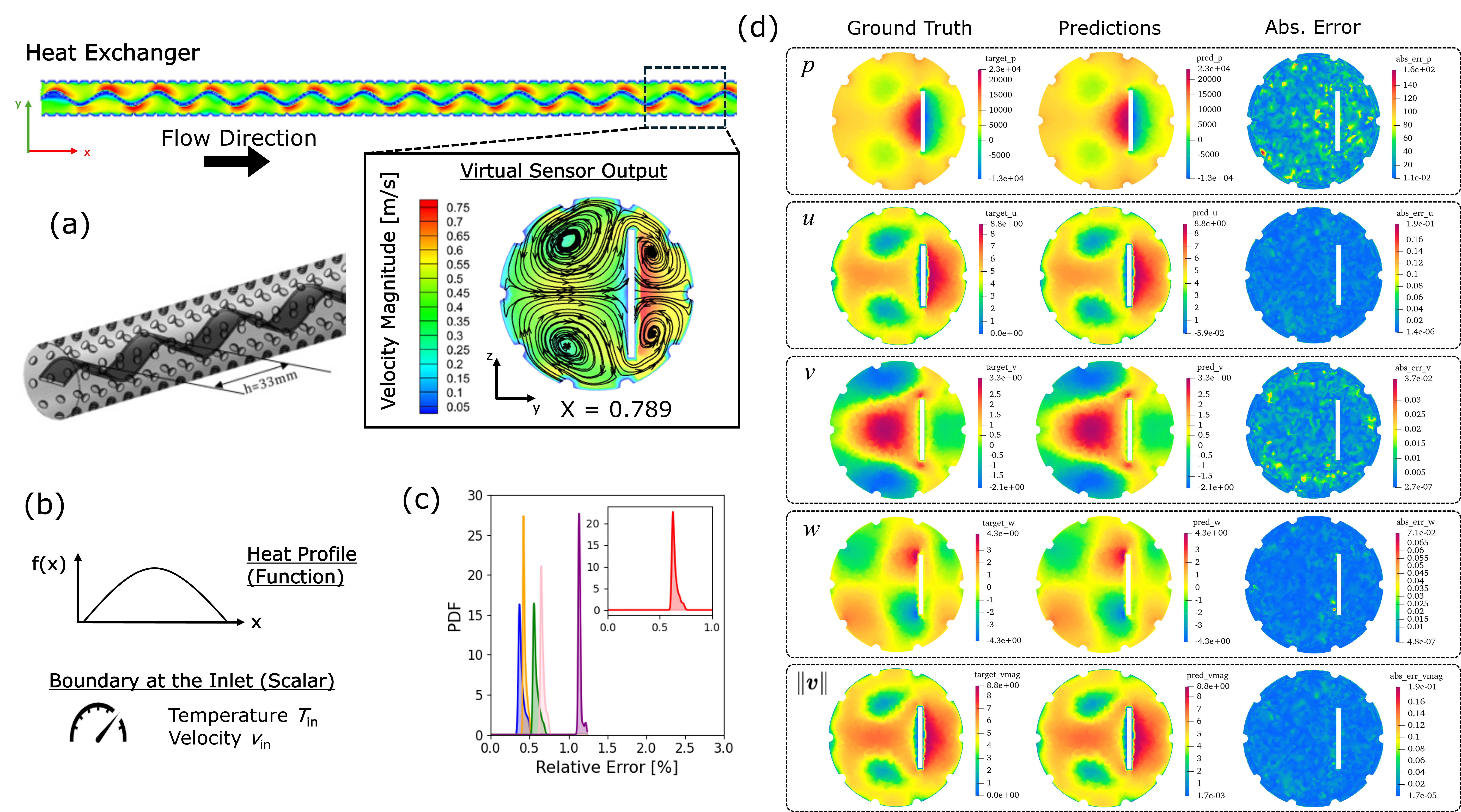}
    \caption{
    \textbf{Heat Exchanger: Problem Formulation and Results.}
    \textbf{(a)} Geometry: wavy tape insert inside a 
    dimpled cylindrical channel. Analysis performed 
    on the 2D axial cross-section at $x=0.789$.
    \textbf{(b)} Boundary inputs: axial heat profile 
    and two scalar inlet conditions, geometrically 
    disjoint from the output cross-section.
    \textbf{(c)} Percentile error distributions for 
    the 10-layer VIRSO across all output fields. 
    Distributions are centered below $1\%$ with 
    narrow spread.
    \textbf{(d)} 50th-percentile reconstruction. 
    VIRSO accurately recovers the primary counter-rotating 
    recirculation cells, secondary eddies, and pressure 
    response to the insert gap. Elevated error is 
    concentrated near the insert junction and 
    channel wall, regions of maximum geometric 
    complexity.
    }
    \label{fig:heat_exchanger}
\end{figure}

The heat exchanger benchmark represents the most 
demanding evaluation of VIRSO, combining the highest 
geometric irregularity, the largest output node count, 
and a four-component coupled output. The geometry 
consists of a dimpled cylindrical channel with a 
wavy tape insert (Figure~\ref{fig:heat_exchanger}a), 
whose surface topology breaks any rotational or 
reflective symmetry present in the subchannel problem. 
The wavy insert generates large recirculation cells 
and complex secondary vortex structures that 
significantly challenge operator learning models 
lacking local resolution capability \cite{AHMED2024104583}.

The output consists of four physical fields at 
3,977 nodes on a two-dimensional axial cross-section 
$\mathcal{Y}_{hx}$: pressure $p(z,y)$ and three 
velocity components $u_x(z,y), u_y(z,y), u_z(z,y)$. 
The boundary input is identical in structure to 
the subchannel problem — two scalar inlet conditions 
and a 100-point axial heat profile — giving a 
reconstruction ratio of:
\begin{equation}
\frac{N \times C}{M} = \frac{3{,}977 \times 4}{102} 
\approx 156{:}1.
\end{equation}
This is the most severely underdetermined configuration 
evaluated in this work. The operator is:
\begin{equation}
\mathcal{G}_{hx}: (\mathbb{R})^2 \times L^2(\mathbb{R}) 
\to (L^2(\mathcal{Y}_{hx}))^4, \quad 
\mathcal{Y}_{hx} \subset \mathbb{R}^2.
\end{equation}

Details regarding model architecture choices are found in Methods \ref{sec:training}.

\textbf{Performance.} As shown in 
Table~\ref{tab:heatexchanger_model_performance}, 
a 14 integral-kernel (\ref{eq:kernel_int}) layer VIRSO achieves a mean $L_2$ error of 
$0.70\%$, outperforming NOMAD ($0.97\%$) and Geo-FNO 
($1.09\%$) at comparable parameter counts, and exceeding 
GNO ($9.40\%$) by more than an order of magnitude. 
A 10 integral-kernel layer model ($0.83\%$) similarly outperforms 
all baselines with fewer parameters than NOMAD or 
Geo-FNO. The velocity magnitude, computed from 
predicted components rather than directly trained, 
shows the lowest reconstruction error among all 
models, confirming that VIRSO maintains physical 
consistency across the velocity field tensor 
without explicit vector-field supervision.

The percentile statistics in Table \ref{tab:heatexchanger_model_percentile} of the Supplementary Material show 
that VIRSO achieves a fully improved sample distribution 
relative to all baselines, from the best-case to the 
95th percentile, with a distribution center of $0.66$--$0.69\%$ 
and a worst-case error of $0.80\%$ — the narrowest 
worst-case bound of any model evaluated. 
The 50th-percentile reconstruction 
(Figure~\ref{fig:heat_exchanger}d) accurately captures 
the two large counter-rotating recirculation cells 
generated by the wavy insert, the opposing spin 
directions of the primary vortices, secondary eddies 
in the component fields, and the pressure response 
to the insert gap. Elevated absolute error is 
concentrated near the insert junction and the 
dimpled channel wall, regions of maximum geometric 
complexity and steepest physical gradients — precisely 
the regions where high graph connectivity is needed, 
as the V-KNN analysis below confirms.

\begin{table}[]
\caption{Model Comparison: Heat Exchanger}
\centering
\begin{tabular}{@{}cccccccc@{}}
\toprule
\multirow{2}{*}{\textbf{Model}} & 
\multirow{2}{*}{\textbf{Parameters}} & 
\multicolumn{6}{c}{\textbf{Mean Relative $L_2$ Error (\%)}} \\ 
\cmidrule(l){3-8} 
& & $p$ & $u_z$ & $u_y$ & $u_x$ & $\|u\|$ & Mean \\ 
\midrule
\textbf{NOMAD}    & 2.69 M & 0.66 & 1.01 & 1.03 & 0.74 & 1.43 & 0.97 \\
\textbf{Geo-FNO}  & 2.70 M & 0.74 & 1.14 & 1.14 & 0.91 & 1.51 & 1.09 \\
\textbf{GNO}      & 0.27 M & 3.37 & 9.89 & 10.75 & 11.15 & 11.82 & 9.40 \\
\textbf{VIRSO$^*$} & 1.66 M & 0.47 & 0.87 & 0.86 & 0.71 & 1.24 & 0.83 \\
\textbf{VIRSO}   & 2.31 M & \textbf{0.40} & \textbf{0.71} & 
\textbf{0.66} & \textbf{0.52} & \textbf{1.18} & \textbf{0.70} \\
\bottomrule
\end{tabular}
\label{tab:heatexchanger_model_performance}
\caption*{\textit{$^*$10-layer VIRSO, 64 modes, function dimension 48.}}
\end{table}

%%% SUBSECTION 4 %%%
\subsection{Spectral-Spatial Collaboration and 
Architecture Analysis}
\label{sec:ablation}

The three benchmarks establish VIRSO's empirical 
performance. We now probe what VIRSO learns and 
why the architecture succeeds, using a 
component-level ablation study on the most 
demanding benchmark. For reference, the removal of a spatial/spectral block includes removing the entire component and collaboration layer while keeping the other block and the identity skip located after the projection mapping $f$.

\textbf{Role of cross-domain input embedding.} 
Unlike standard operator benchmarks in which the 
input and output domains coincide, VIRSO's nuclear 
use cases involve a geometrically disjoint input 
space: the boundary observations are defined on 
the axial direction while the output fields are 
defined on the transverse cross-section. This 
requires that the boundary input be lifted into 
a latent representation that is then broadcast 
to all output nodes, rather than evaluated 
point-wise at their coordinates. To quantify 
the necessity of this design choice, 
our analysis in Table \ref{tab:model_component_comparison} in the Supplementary Material
isolates the FCN latent embedding and the 
residual skip connection in the 10-layer 
heat exchanger model. Removing both components 
raises the mean error from $0.83\%$ to $4.16\%$, 
a five-fold degradation. The embedding alone 
allows for a $4.16\% \to 0.93\%$ reduction, consistent with the 
interpretation that the FCN extracts the 
physically relevant low-dimensional structure 
of the heat profile — primarily its amplitude 
$A$ and spatial extent $H$ — from the 
100-dimensional input representation. Moreover, the addition of a latent embedding can include sequential-based models that can handle varying histories or profile resolutions without retraining. This is demonstrated with the Lid-Driven Cavity implementation of VIRSO (Methods \ref{sec:training}).

\textbf{What VIRSO's spectral-spatial collaboration 
learns.} Further analysis in Table \ref{tab:spectral_spatial_results} of the Supplementary Material
decomposes the 10-layer VIRSO model into its 
spectral-only, spatial-only, and combined forms. 
Three findings emerge. First, the spectral-only 
model with residual skip connections ($0.90\%$) 
substantially outperforms all external benchmarks 
($0.97$--$9.40\%$), demonstrating that graph 
spectral convolution is a primary learning mechanism 
and that it alone can be sufficient to solve the inverse 
reconstruction problem at competitive accuracy in the presence of negligible high-frequency physics. 
Second, removing all residual skip connections from 
the spectral-only model raises the error from $0.90\%$ 
to $5.76\%$, revealing that the skip connections 
serve as local high-frequency feature preservers: 
the 64 spectral eigenmodes capture the dominant 
low-frequency physical structure of the heat exchanger 
flow, but the residual paths preserve the high-frequency 
modes beyond the truncation threshold that encode 
wall shear and vortex detail. Moreover, with the combined spatial-spectral analysis, we found in Table \ref{tab:model_component_comparison} of the Supplementary Material that the addition of the skip connection after the collaboration layer showed consistent improvement in performance (4.16\% to 1.32\% from adding skip without embedding and 0.93\% to 0.83\% from utilizing the skip connection after adding the embedding FCN), reiterating the importance of residual-based learning for gradient flow stability and high-frequency mode preservation within neural operator frameworks. Third, the spatial 
aggregation component reduces the combined model 
error from $0.90\%$ to $0.83\%$, a modest but 
consistent improvement that reflects local 
fine-tuning of the spectral prediction rather 
than independent learning — confirmed by the 
spatial-only model's catastrophic failure 
at $16.27\%$. Together, these findings establish 
that VIRSO's spectral-spatial collaboration is 
not architecturally redundant: spectral convolution 
handles global physical consistency, skip connections 
preserve high-frequency local structure, and spatial 
aggregation provides calibration at boundaries and 
irregular geometric features. For more complex and large-scale geometries, we believe that the spatial analysis would provide a larger contribution towards high-frequency information that cannot be effectively captured by only residual skips and spectral-based learning. 

%%% SUBSECTION 5 %%%
\subsection{Graph Construction: 
The Variable-KNN Principle}

\begin{figure}[htbp]
    \centering
    \includegraphics[width=0.95\textwidth]{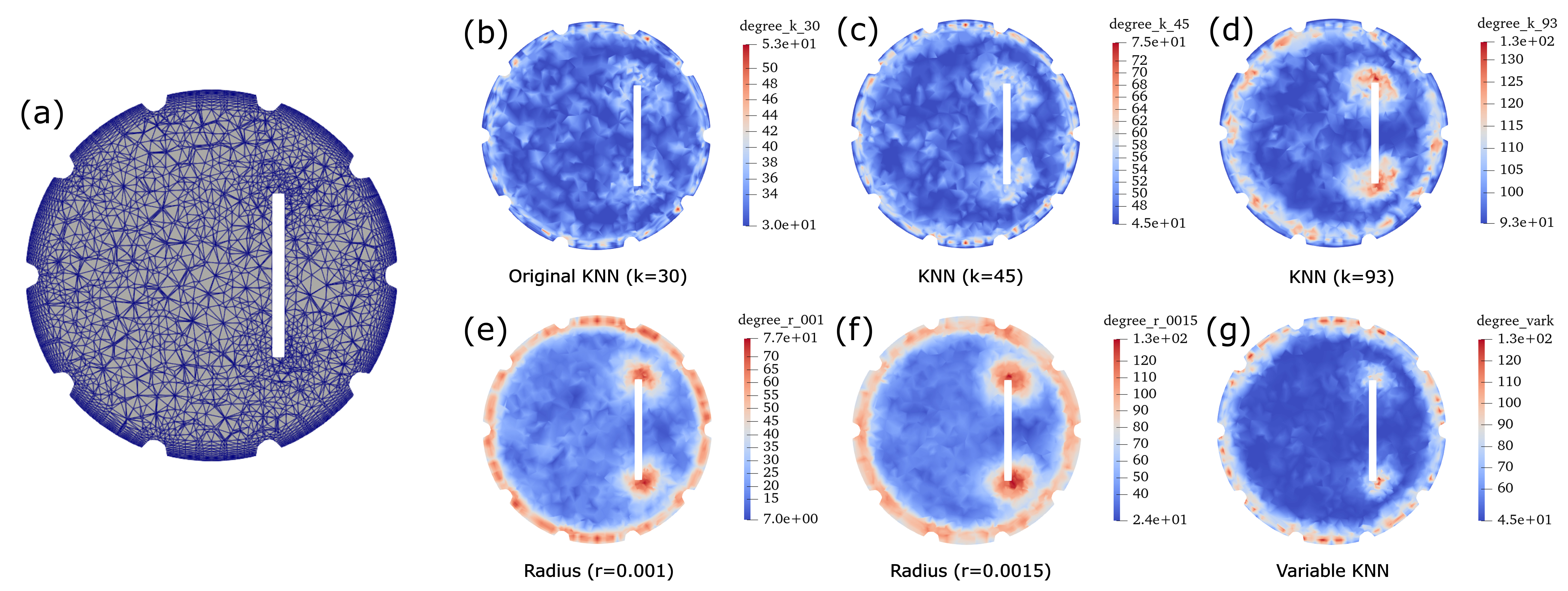}
    \caption{
    \textbf{Graph Topology and Node Degree Distribution.}
    \textbf{(a)} Mesh geometry of the heat exchanger, 
    adapted to the VIRSO graph node configuration. 
    Higher mesh density is visible near the insert 
    junction and channel walls, corresponding to 
    regions of steepest physical gradients.
    \textbf{(b,c)} Degree distribution for KNN with 
    $k=30$ and $k=45$: modest variation with slightly 
    elevated connectivity near high-density regions, 
    but insufficient targeting.
    \textbf{(d)} KNN with $k=93$: high-density regions 
    receive elevated degree, improving performance, 
    but the enforced minimum of 93 throughout the 
    interior introduces significant inefficiency.
    \textbf{(e)} Radius with $r=0.001$: naturally 
    concentrates high degree near geometrically 
    complex regions; outperforms KNN with $k=30,45$.
    \textbf{(f)} Radius with $r=0.0015$: matches the 
    maximum degree of KNN with $k=93$ near walls 
    and junctions but maintains a minimum of only 24 
    in the interior, achieving superior performance 
    with fewer edges.
    \textbf{(g)} Variable-KNN (V-KNN): explicitly 
    assigns connectivity proportional to local mesh 
    density. Achieves the same maximum degree as 
    (d) and (f) near geometric complexity, maintains 
    minimum degree 45 in the interior, and uses 
    only 270K edges — the most efficient configuration 
    achieving the best overall accuracy.
    }
    \label{fig:graph_comparison}
\end{figure}

Graph topology constitutes an underexplored source 
of inductive bias in graph-based operator learning. 
The standard KNN graph assigns a uniform minimum 
degree $k$ to all nodes regardless of their geometric 
context, which is efficient but physically naive. 
In finite element analysis, mesh refinement 
concentrates resolution near geometrically complex 
regions — walls, corners, and high-curvature 
surfaces — precisely where physical gradients 
are steepest. We hypothesize that graph connectivity 
in neural operators should obey the same principle: 
nodes in high mesh-density regions, which correspond 
to regions of high geometric complexity, require 
higher neighbor counts to propagate physical 
information across the steep local gradients 
that arise there. The radius graph provides 
a natural test of this hypothesis, since it 
assigns connectivity proportional to local node density. It should be noted that previous concerns about graph regularity and the effect of node hubs towards over-smoothing \cite{Vega-Oliveros_2014, HUANG2023110556} has not been addressed but most-likely has little effect in the performance of VIRSO due to its combined spatial and spectral architecture and the nature of the regression problems utilized, where variation in flow complexity might require differing node degrees. 

Table \ref{tab:radius_knn_comparison} of the Supplementary Material confirms the 
hypothesis quantitatively with the 10-layer VIRSO model. Among KNN graphs, increasing 
$k$ uniformly from 30 to 93 reduces the mean error 
from $0.83\%$ to $0.67\%$, but at a cost of tripling 
the edge count. Among radius graphs, $r=0.001$ 
outperforms KNN with $k=45$ ($0.71\%$ vs. $0.79\%$) 
despite fewer edges, because the radius graph 
preferentially concentrates high connectivity near 
the wall and insert junction — the regions shown 
in Figure~\ref{fig:graph_comparison}a to have the 
highest mesh density. Radius with $r=0.0015$ 
outperforms KNN with $k=93$ ($0.62\%$ vs. $0.67\%$) 
while using 284K fewer edges, confirming that 
targeted high-degree assignment to geometrically 
complex regions is more effective than uniform 
high-degree assignment throughout the domain.

These results motivate Variable-KNN (V-KNN), 
a mesh-density-adaptive graph construction 
strategy that assigns neighbor counts proportional 
to local node density, estimated from an initial 
radius graph at radius $5 \times 10^{-4}$. 
Each node's degree is set to a fraction of 
$k_{\max}$ equal to its normalized local density, 
with a minimum of $k_{\min}$ enforced throughout 
the interior. With $k_{\max}=134$ and $k_{\min}=45$, 
V-KNN produces a graph with 270K edges — more 
efficient than both KNN with $k=93$ (408K) and 
radius with $r=0.0015$ (292K) — that achieves 
a mean error of $0.59\%$ (in Table \ref{tab:new_graph_results} of the Supplementary Material), 
the best result across all graph construction 
strategies tested. Figure~\ref{fig:graph_comparison}g 
confirms that V-KNN successfully concentrates 
high connectivity near the dimpled channel wall 
and wavy-insert junction, while maintaining a 
controlled minimum degree of 45 in the interior. Beyond reconstruction accuracy, V-KNN provides a hardware efficiency advantage directly relevant to edge deployment. With 270K edges versus 408K for KNN at $k=93$ — the uniform configuration that most closely approaches V-KNN's accuracy — V-KNN reduces the total edge count by approximately 34\% ($\Delta = 138\mathrm{K}$ edges). In the spatial aggregation block, inference energy and latency scale approximately linearly with edge count, because each edge corresponds to an independent gather-accumulate operation. V-KNN therefore simultaneously improves spectral eigenmode alignment, reconstruction accuracy, and memory access efficiency — constituting a hardware-aware graph construction principle rather than a purely topological one, applicable to any graph-based operator learning system on irregular physical domains.

To understand the mechanism by which V-KNN 
improves performance, 
Table \ref{tab:spectral_spatial_results_new_graph} of the Supplementary Material repeats the spectral-spatial decomposition on 
the V-KNN graph. V-KNN improves the spectral-only 
model with skip connections from $0.90\%$ to $0.71\%$ 
— a significant gain — while degrading both the 
spatial-only model and the spectral model without 
skip connections. This dissociation reveals 
the mechanism: V-KNN restructures the graph 
Laplacian's eigenmodes to better align with the 
physical structure of the heat exchanger flow, 
shifting energy toward higher-frequency modes 
that capture wall shear and vortex detail. 
This eigenmode restructuring benefits the 
spectral operator, which explicitly decomposes 
the signal into graph spectral modes, but 
penalizes the spatial aggregation operator, 
which lacks the global spectral view needed 
to correctly interpret the redistributed connectivity.

%%% SUBSECTION 6 %%%
\subsection{Computational Efficiency and 
Energy Consumption}

The computational analysis presented here serves a specific and bounded purpose: establishing that VIRSO satisfies the deployment feasibility requirements of edge-constrained real-time sensing, not claiming architectural superiority in energy consumption over all possible model classes. This distinction is important. GPU-level power draw and energy-per-inference measurements characterize the hardware execution cost of a given model on a given accelerator; they do not constitute an architecture-level efficiency comparison unless evaluated under identical hardware conditions across architectures, which the cross-architecture analysis in Table 4 and Table 5 provides. The hardware-level measurements reported for the GNO, Geo-FNO, NOMAD, and VIRSO architectures are obtained on identical hardware under identical workload conditions, providing a valid architecture-level comparison within this specific benchmark. The Jetson Orin Nano measurements reported in Table 6 additionally establish hardware portability that the pretrained VIRSO model executes within a single-digit watt continuous power envelope on embedded hardware without retraining, quantization, or architectural modification, which is a necessary condition for edge deployment in I\&C systems, not merely an engineering convenience. The governing constraint for inclusion in safety-critical embedded sensing systems is not minimum energy in isolation but rather the joint satisfaction of accuracy, latency, and power requirements; the analysis below evaluates all three simultaneously and identifies the VIRSO configuration that lies on the Pareto frontier of this constraint space for each deployment target.

Deployment of neural operators for real-time 
virtual sensing in nuclear systems requires 
not only predictive accuracy but energy-efficient 
inference, particularly under edge computing 
constraints where total I\&C power budgets 
are limited~\cite{10545889, ALSHARIF20251739}. 
Prior neural operator work does not characterize inference under hardware deployment constraints. The analysis below is, to our knowledge, the first systematic hardware efficiency characterization of a neural operator under the compute-versus-memory-bandwidth hierarchy of its target deployment platforms, and the first to validate operator-based inference end-to-end on an embedded edge accelerator. We report device-level GPU inference statistics for the 310-sample heat exchanger test set on an NVIDIA H200 GPU and board-level embedded inference on an NVIDIA Jetson Orin Nano (Section~\ref{sec:edge_deployment}), enabling a controlled comparison across two memory-subsystem regimes. We report GPU inference statistics for the 310-sample heat exchanger test set on an NVIDIA H200 GPU, using NVIDIA built-in profiling software to measure kernel utilization, memory allocation, instantaneous power, latency, and per-iteration energy consumption (Table~\ref{tab:initial_energy_results}).

\begin{table}[H]
\caption{Inference Energy and Memory: Heat Exchanger}
\centering
\begin{tabular}{@{}ccccccccc@{}}
\toprule
\textbf{Model} & \textbf{Param.} & \textbf{FLOPs} & \textbf{GPU\%} &
\textbf{Mem\%} & \textbf{Mem (MiB)} & \textbf{Pwr (W)} & 
\textbf{Lat. (ms)} & \textbf{Energy (J/it)} \\
\midrule
\textbf{NOMAD}     & 2.69 M & 11.29 G & 30.92 & 1.12 & 999.92 & 
196.03 & 2.35 & 0.41 \\
\textbf{Geo-FNO}   & 2.70 M & 1.58 G & 23.63 & 2.02 & 1059.39 & 
139.39 & 4.94 & 0.59 \\
\textbf{GNO}       & 0.27 M & 429.49 G & 86.78 & 36.15 & 1965.99 & 
572.00 & 20.48 & 10.07 \\
\textbf{VIRSO}     & 1.66 M & 2.03 G & 42.35 & 5.71 & 1039.98 & 
193.35 & 7.77 & 1.30 \\
\textbf{VIRSO (Spatial)}      & 0.11 M & 0.99 G & 32.77 & 6.95 & 
1031.65 & 178.15 & 8.93 & 1.40 \\
\textbf{VIRSO (Spectral)}     & 1.58 M & 0.98 G & 17.34 & $<$1 & 
937.93 & 124.41 & 8.18 & 0.86 \\
\textbf{VIRSO (V-KNN)} & 1.66 M & 2.56 G & 64.11 & 14.93 & 1153.94 & 
265.02 & 8.32 & 1.91 \\
\bottomrule
\end{tabular}
\label{tab:initial_energy_results}
\caption*{\textit{Spatial and Spectral refer to Spatial-Only and Spectral-Only VIRSO.}}
\end{table}

\textbf{Memory scalability.} All models except 
GNO allocate approximately 1 GiB of device memory. 
GNO requires nearly 2 GiB — a consequence of 
its dense message-passing scheme — and is 
memory-limited in all three benchmarks. VIRSO 
uses approximately 0.1 GiB more than NOMAD 
and Geo-FNO, reflecting the additional graph 
spectral computation, while remaining well within 
standard GPU memory constraints.

\textbf{Latency.} From the VIRSO model implementations explored, we found a real-time latency ranging from about 4.29 - 10.32 ms for the Heat Exchanger dataset of 3,977 evaluation nodes. The sequential aspect of VIRSO is responsible for the latency in model inference. This is demonstrated by Table \ref{tab:light_model_comparisons} of the Supplementary material with a 14-layer VIRSO model requiring approximately three times the latency for predictions than a 2-layer lightweight model with width 48 and mode count of 40. Compared to other neural operators, VIRSO only requires slightly more time for predictions with latency statistics within the same order of magnitude as NOMAD and Geo-FNO. Such computational requirements are a drastic improvement to previous graph methods that are either unable to scale to our chosen benchmarks or have high computation requirements such as GNO (Table \ref{tab:initial_energy_results}). 

\textbf{Comparison with ANSYS Fluent.} For the Heat Exchanger and Subchannel, our 2D slice results originate from full 3D simulation. To provide an estimate of the amortized inference speed up of VIRSO compared to Fluent, we must observe the Lid-Driven Cavity which was purely two dimensional with 4,225 evaluation nodes. ANSYS Fluent required approximately 36 minutes to generate a solution on an AMD EPYC 7763 (``Milan'') CPU while our VIRSO model configured for the LDC use case required around 88 milliseconds on an NVIDIA H200 GPU which corresponds to a speedup of more than 4 orders of magnitude. VIRSO is able to improve upon the high computational requirement of preceding graph methodologies and provide highly-accurate irregular field reconstruction in real-time.

\textbf{Power Draw.} In Table \ref{tab:initial_energy_results}, we also found that VIRSO without V-KNN required similar power draw to NOMAD and slightly higher draw (within 60W) than Geo-FNO. The spectral only model required lower instantaneous power than the other operator models, emphasizing the improved computational efficiency of VIRSO compared to previous graph operators such as GNO which required approximately half a kilowatt of power.

\textbf{Energy cost of spatial aggregation.} 
The primary energy cost of VIRSO relative to 
NOMAD and Geo-FNO is the local spatial graph 
convolution. Table~\ref{tab:initial_energy_results} 
also isolates this cost: the spatial-only component 
requires 1.40 J/it versus 0.86 J/it for the 
spectral-only component with skip connections. 
This is a direct consequence of the 
message-passing implementation in PyG \cite{PyG}, 
where energy consumption scales with 
total edge aggregation operations — 
approximately linear in the product of 
node count and average degree. The combined 
VIRSO model (1.30 J/it) falls between its 
components because the collaboration 
mechanism partially offloads learning 
to the spectral path, reducing the 
effective contribution of the spatial path.

\textbf{Performance trade-off.} 
VIRSO consumes 3--5 times more energy 
per inference than NOMAD or Geo-FNO at 
their full parameter counts 
(Table~\ref{tab:initial_energy_results}), 
while achieving approximately 20--30\% 
lower mean error. For applications where 
energy budget is the primary constraint,
the spectral-only VIRSO configuration 
(0.86 J/it) provides the best trade-off: 
it consumes only twice the energy 
of NOMAD while outperforming all external 
baselines on the heat exchanger benchmark. Accuracy-critical scenarios with high-frequency physics would then required the inclusion of the spatial block.

\textbf{Energy-delay product.} To jointly characterize the latency and energy trade-off, we compute the energy-delay product $\mathrm{EDP} = E_{\mathrm{iter}} \times t_{\mathrm{lat}}$, which penalizes configurations that sacrifice latency for energy savings or vice versa. From Table~\ref{tab:initial_energy_results}, GNO incurs an EDP of $10.07 \times 20.48 \approx 206$\,J$\cdot$ms — more than $20\times$ higher than the full VIRSO configuration ($1.30 \times 7.77 \approx 10.1$\,J$\cdot$ms). The spectral-only VIRSO achieves the most favorable EDP among graph-based operators ($0.86 \times 8.18 \approx 7.0$\,J$\cdot$ms), and approaches Geo-FNO ($0.59 \times 4.94 \approx 2.9$\,J$\cdot$ms) and NOMAD ($0.41 \times 2.35 \approx 0.96$\,J$\cdot$ms) while delivering substantially lower reconstruction error on the heat exchanger benchmark.

\textbf{Power-normalized accuracy.} We define a power-normalized accuracy metric $\eta = (100\,/\,\bar{e}_{L_2}[\%])\,/\,P[\mathrm{W}]$, quantifying reconstruction accuracy delivered per watt of instantaneous GPU power. The spectral-only VIRSO achieves $\eta = (100\,/\,0.90)\,/\,124.41 \approx 0.89\;\%^{-1}\mathrm{W}^{-1}$ and full VIRSO configuration achieves $\eta = (100\,/\,0.83)\,/\,193.35 \approx 0.62\;\%^{-1}\mathrm{W}^{-1}$, compared to $\eta = (100\,/\,0.97)\,/\,196.03 \approx 0.53\;\%^{-1}\mathrm{W}^{-1}$ for NOMAD. Despite requiring more energy per inference in absolute terms, the spectral-only and full VIRSO delivers superior reconstruction accuracy per unit of power consumption. These values reflect device-level GPU measurements on the H200 and are not directly comparable to the board-level Jetson measurements in Table~\ref{tab:jetson_nano}.

A 2-layer lightweight VIRSO (0.26 M parameters, 
0.54 J/it, width 48, and mode count of 40) achieves comparable energy 
consumption to the original Geo-FNO and NOMAD and their lightweight versions while maintaining solid performance compared to high performance degradation of the lightweight versions of Geo-FNO and NOMAD
(Table~\ref{tab:pareto_summary} and 
Table \ref{tab:light_model_comparisons} and \ref{tab:light_model_comparisons_energy} of the Supplementary Material). 

In terms of latency, lower number of layers reduces prediction time, with the 2-layer model (4.29 ms) providing improved latency comparable to other operators without extreme performance degradation such as NOMAD's transition to lightweight (Table~\ref{tab:pareto_summary}, 
Table \ref{tab:light_model_comparisons}, and \ref{tab:light_model_comparisons_energy} of the Supplementary Material).

The energy analysis establishes that the 
appropriate VIRSO configuration is task-dependent: 
the full spectral-spatial model is preferred 
when accuracy is the governing constraint, 
the spectral-only model when energy efficiency 
is primary, and the lightweight model when 
both constraints are simultaneously binding.

For accuracy-critical applications where computational resources are unconstrained, the full spectral-spatial VIRSO with V-KNN (1.91 J/it) is preferred. For edge-deployed sensing where total I\&C power budgets impose strict limits, the spectral-only configuration (0.86 J/it) provides the best accuracy-energy trade-off while outperforming all baselines but risking the loss of higher frequency analysis and calibration with the spatial block. For the most resource-constrained settings, the 2-layer lightweight model (0.54 J/it) maintains reconstruction errors below 2\%, substantially better than comparably sized alternatives, at energy consumption comparable to conventional operator methods. Lastly, latency-constrained applications within an Nvidia H200 device would require the lowest layer count for VIRSO, such as the 2-layer version (4.29 ms). In addition, graph construction and degree count are other considerations with higher neighbors, or essentially increased aggregation, typically resulting in higher power draw for the combined spatial-spectral analysis (shown with V-KNN in Table \ref{tab:initial_energy_results}).

Table~\ref{tab:pareto_summary} consolidates accuracy and hardware efficiency across all evaluated configurations on the heat exchanger benchmark, enabling direct comparison across the accuracy--energy trade-off space. The spectral-only VIRSO occupies the Pareto frontier among graph-based operators: it achieves lower reconstruction error than all baselines while requiring 0.86\,J/it and 8.18\,ms latency. The 2-layer VIRSO provides NOMAD-comparable energy consumption (0.54\,J/it) and latency (4.29\,ms) at substantially lower reconstruction error than any lightweight baseline (1.95\% vs.\ 4.21--8.24\%), establishing VIRSO as the preferred architecture across the full range of accuracy-constrained deployment conditions.

\begin{table}[H]
\caption{Accuracy--Efficiency Summary: Heat Exchanger (H200, device-level GPU measurements)}
\centering
\begin{tabular}{@{}lccccc@{}}
\toprule
\textbf{Model} & \textbf{Mean $L_2$ (\%)} & \textbf{FLOPs} & \textbf{Energy (J/it)} & \textbf{Lat.\ (ms)} & \textbf{EDP (J$\cdot$ms)} \\
\midrule
GNO                        & 9.40 & 429.49 G & 10.07 & 20.48 & 206.2 \\
Geo-FNO (full, 2.70\,M)   & 1.09 & 1.58 G & 0.59  & 4.94  & 2.91  \\
Geo-FNO (light, 0.26\,M)  & 4.21 &  1.16 G & 0.56  & 5.10  & 2.86  \\
NOMAD (full, 2.69\,M)     & 0.97 & 11.29 G & 0.41  & 2.35  & 0.96  \\
NOMAD (light, 0.26\,M)    & 8.24 & 1.28 G & 0.23  & 2.00  & 0.46  \\
\midrule
VIRSO (2-layer, 0.26\,M)         & 1.95 & 0.61 G & 0.54 & 4.29 & 2.32 \\
VIRSO (spectral-only, 1.58\,M)   & 0.90 & 0.98 G & 0.86 & 8.18 & 7.03 \\
VIRSO (10-layer full, 1.66\,M)   & 0.83 & 2.03 G & 1.30 & 7.77 & 10.1 \\
\bottomrule
\end{tabular}
\label{tab:pareto_summary}
\caption*{\textit{EDP = Energy $\times$ Latency. Lightweight variants use ${\approx}0.26$\,M parameters (Tables ~\ref{tab:pareto_summary} and \ref{tab:light_model_comparisons} from Supplementary Material). VIRSO (10-layer full) accuracy is the 1.66\,M configuration (VIRSO$^*$, Table~\ref{tab:heatexchanger_model_performance}); hardware numbers are from Table~\ref{tab:initial_energy_results}. All values are H200 device-level GPU measurements and are not directly comparable to Jetson board-level measurements in Table~\ref{tab:jetson_nano}.}}
\end{table}

\subsection{Towards Edge Deployment: Embedded Inference on Resource-Constrained Hardware}
\label{sec:edge_deployment}
Unlike conventional surrogate models evaluated post hoc, VIRSO is designed such that its operator decomposition aligns with hardware execution constraints, enabling it to function as a physically realizable sensing instrument under edge conditions.

\begin{table}[H]
\caption{Embedded inference performance of pretrained 10-layer (full and spectral-only) and the 2-layer VIRSO model.}
\centering
\begin{tabular}{@{}cccccccc@{}}
\toprule
\textbf{Model} & \textbf{Param.} & \textbf{FLOPs} & \textbf{Avg. Lat. (ms/it)} & 
\textbf{Avg. Pwr (W)} & \textbf{Energy (J/it)} & \textbf{Peak RAM (GB)} & 
\textbf{Mean L2 (\%)}\\
\midrule
10-L (Full) & 1.66 M & 2.04 G & 562.92 & 7.58 & 4.72 & 5.16 & 0.84\\
\midrule
10-L (Spec.) & 1.58 M & 0.98 G & 58.77 & 7.06 & 0.84 & 5.14 & 0.91\\
\midrule
2-L & 0.26 M & 0.61 G & 104.03 & 7.51 & 1.25 & 5.32 & 1.96\\
\bottomrule
\end{tabular}
\label{tab:jetson_nano}
\caption*{\textit{N-L represents N spectral-spatial layers for VIRSO, FLOPs per inference sample are computed via \texttt{torch.profiler} on the 3,977-node Heat Exchanger test set.}}
\end{table}

The preceding analysis characterizes the energy--accuracy trade-off of VIRSO on datacenter-scale GPUs. A related practical question is whether the same pretrained operator is hardware-portable and can also be executed on embedded hardware, where memory, board-level power, and thermal margins are substantially more limited. This question is relevant to virtual sensing deployments in which inference may need to be performed near the sensing hardware rather than on remote accelerator infrastructure.

\textbf{Experimental setup.}
We evaluated the three pretrained VIRSO models, 10-Layer Full, 10-Layer Spectral-Only, 2-Layer Full, on an NVIDIA Jetson Orin Nano (8\,GB), an embedded platform with an Ampere-based GPU and shared system memory. Our analysis is deployed towards the Heat Exchanger dataset with the same 3,977-node output. The 10-Layer models utilize a width and mode count of 48 and 64 while the 2-Layer model has the same width but only 40 spectral modes. All three models are trained on a $k=30$ KNN graph. No retraining, quantization, pruning, or architecture-specific modification was introduced. Inference was performed on the full 310-sample Heat Exchanger test set, and reported values correspond to the mean over three independent runs with board-level telemetry recorded throughout execution. A summary of the results is shown in Table \ref{tab:jetson_nano} where we display average latency, power, energy per iteration, peak RAM usage, and the mean relative L2 error over all multiple outputs in the Heat Exchanger dataset.

\textbf{Latency.}
Across the three runs, the 10-layer VIRSO achieved an average latency of $562.92$\,ms/iteration (1.78\,samples/s) for the full version and $58.77$\,ms/iteration (17.0\,samples/s) for the spectral-only version (Table~\ref{tab:jetson_nano}). The 2-layer model achieved an average latency of $104.03$\,ms/iteration (9.61\,samples/s). Although the resource-constrained operation of the Jetson Nano results in a significantly higher latency than the NVIDIA H200, VIRSO's efficient graph analysis allows for a sub-second per-sample regime on embedded hardware without deployment-specific simplification. Future acceleration is required to push latency, especially for the full 10-layer model, to realistic real-time conditions (<100ms). An unexpected trend in the latency results is that the 10 layer spectral-only latency is lower than the 2-layer model. Compared to an NVIDIA H200, the Jetson Nano struggles significantly more on the point-wise aggregation within the spatial block, making such local analysis the main bottleneck for latency instead of layer count which was observed with the H200 analysis. It is most likely that the sophisticated optimization of the H200, compared to the resource-constrained Nano, is able to better handle the local aggregation algorithms implemented by the python-based PyG library \cite{PyG}. Quantitatively, the full 10-layer model is $562.92\,/\,58.77 \approx 9.6\times$ slower than the spectral-only configuration on the Jetson Nano, whereas the corresponding latency difference on the H200 is less than 0.5\,ms (Table~\ref{tab:initial_energy_results}). This hardware-dependent dissociation reflects a fundamental difference in memory-subsystem architecture: graph message passing is a \emph{memory-bandwidth-bound} operation whose irregular scatter-gather pattern incurs disproportionate cost on the Jetson Orin Nano's unified LPDDR5 memory, whereas spectral convolution reduces to dense matrix multiplications that are compute-bound and therefore hardware-portable across the two platforms. The spectral-only configuration eliminates this bandwidth bottleneck entirely, delivering 17.0\,samples/s on the Jetson Nano without retraining or architectural modification.

\textbf{Power and energy.}
The average board-level power during inference for all models ranged from $7.06$\,W (spectral-only configuration) to $7.58$\,W (full 10-layer), measured as the mean \texttt{VDD\_IN} rail power integrated over the full 310-sample evaluation window using \texttt{tegrastats} at 20\,ms sampling intervals. This board-level measurement encompasses GPU, CPU, and shared-memory activity throughout inference. All configurations sustain $<$10\,W total board power with only slight dependence on model architecture or layer count, confirming operational compatibility with strict embedded power envelopes. With the Jetson Nano, we are able to address the high power consumption seen with the H200 in Table \ref{tab:initial_energy_results} on the NVIDIA H200. In other words, the main bottleneck for the real-time performance on the Jetson Nano, within the context of the benchmarks presented, is model latency which is emphasized in the energy consumption results. The full 10-layer VIRSO model achieved $4.721$ J/iteration due to its higher latency while the spectral only was less than one joule. These values are not directly comparable to the datacenter GPU measurements in Table~\ref{tab:initial_energy_results}, because the telemetry domains differ (board-level on Jetson versus device-level on H200). Nevertheless, the observed board power remained within a single-digit watt range throughout execution.

\textbf{Resource stability.}
Resource usage was stable across runs. Peak RAM usage and observed temperature for all models was no higher than 5.32\,GB and 47.93$^\circ$C (maximum for both is the full 10-layer). The lowest Peak RAM was 5.14\,GB and the lowest temperature was 44.56$^\circ$C (both for the full 2-layer). GPU utilization over all three models averaged from 17.65\% (2L) to 32.95\% (10L full) and peaked from 67\% (2L) to 79\% (both 10L models). Together, these measurements indicate that the full inference pipeline completed on the embedded platform without memory exhaustion or thermal instability.

\textbf{Performance trade-off.} 
As expected, the reconstruction performance of all three models in Table \ref{tab:jetson_nano} is almost identical to the performance on the NVIDIA H200. As a result, the energy-performance trade-off is similar. For accuracy-critical applications, the full 10-Layer model is preferred while the spectral-only version is preferred for energy-constrained and latency-constrained conditions. For ideal real-time performance on the Jetson Nano, latency is the main concern, and the spectral-only version of VIRSO provides a desired <100ms latency. It should be noted that excluding the use of the spatial block should be avoided, especially with large-scale, complex applications where the local aggregation can have a larger contribution to reconstruction error. Further work within edge computing devices such as the Jetson Nano should address accelerating VIRSO to the desired latency standards. We further quantify computational complexity in terms of floating-point operations (FLOPs), demonstrating that VIRSO achieves lower algorithmic cost while reducing memory-bound operations through localized spatial computation.

Critically, the ablation analysis (Section \ref{sec:ablation}) and energy analysis above reveal a hardware-architecture correspondence that is, to our knowledge, uncharacteristic in prior operator learning work: the spectral path's dominant operations are dense matrix multiplications (compute-bound), while the spatial path's dominant operations are irregular edge gather-scatter (memory-bandwidth-bound). This dissociation directly motivates the deployment-aware architectural choice of removing the spatial block and presents VIRSO as a flexible architecture, unique among other operators. In other words, when encountering strict computational constraints and negligible high-frequency physics, the spectral-only configuration can be utilized, positioning VIRSO as a hardware-aligned design that can be uniquely configured to specific edge-compute applications. When complex and higher frequency physics is no longer negligible, the spatial portion is then required and provides a calibration for improved accuracy and sophisticated irregularity-handling with increased computation that still is far below previous graph methodologies and comparable to existing solutions. Further work is necessary for hardware-based acceleration of the entire VIRSO model, but we present VIRSO's initial feasibility for edge deployment.

%%% SUMMARY BOX %%%
\noindent
\vspace{2mm}
\begin{tcolorbox}[
    colback=gray!10!white,
    colframe=blue!50!black,
    title=\textbf{Key Findings from VIRSO Evaluation},
    coltitle=white,
    fonttitle=\bfseries,
    left=0mm,
    right=0mm,
    boxsep=1mm,
    arc=1mm,
    outer arc=1mm
]
\begin{itemize}
    \item \textbf{Consistent accuracy in real-time under severe 
    underdetermination:} VIRSO achieves mean relative 
    $L_2$ errors below $1\%$ across all three benchmarks, 
    at reconstruction ratios ranging from $47{:}1$ to 
    $156{:}1$, with fewer parameters than competing 
    operator architectures in each case. Compared to traditional solvers, VIRSO provides significantly faster inference with a speed of more than 4 orders of magnitude with the Lid-Driven Cavity.
    
    \item \textbf{Spectral convolution is a primary 
    learning mechanism:} A spectral-only VIRSO with 
    graph Laplacian eigenmodes to better align 
    with the physical structure of the domain and residual skip connections for local preservation outperforms all external 
    baselines, establishing that graph spectral 
    decomposition captures the dominant physical 
    structure of the coupled multiphysics field. 
    Spatial aggregation provides local calibration, 
    not independent learning.
    
    \item \textbf{Residual skip connections preserve 
    high-frequency physics:} Removing skip connections 
    from the spectral model raises error from $0.90\%$ 
    to $5.76\%$, demonstrating that physical information 
    at spatial frequencies beyond the spectral truncation 
    threshold is retained through the residual paths 
    rather than through spectral modes.
    
\item \textbf{Graph topology is a physical 
    inductive bias:} V-KNN, which concentrates 
    graph connectivity near high mesh-density regions 
    by analogy with adaptive mesh refinement, 
    achieves the best performance ($0.58\%$ mean error) 
    at lower edge count than uniform high-$k$ alternatives. 
    This improvement operates by restructuring the graph 
    Laplacian eigenmodes to align with the physical flow 
    structure, and simultaneously reduces edge count by 
    34\% relative to the uniform high-$k$ configuration 
    that matches its accuracy, lowering memory bandwidth 
    demand and gather-accumulate operations per inference cycle.
    
    \item \textbf{Configurable energy-accuracy 
    trade-off:} The spectral-only VIRSO variant 
    (0.86 J/it) provides a deployable balance 
    between energy efficiency and predictive accuracy, 
    consuming less than twice the energy of NOMAD 
    while outperforming all baselines. A 2-layer 
    lightweight configuration achieves NOMAD-comparable 
    energy consumption and latency with substantially lower 
    reconstruction error to other similarly constrained operators.

\item \textbf{Edge-Deployment Feasiblity, Hardware Portability, and Architecture Flexibility:} VIRSO is, to our knowledge, the first neural operator whose internal decomposition is utilized to address the compute-constraints of edge device applications. We present its hardware-portability with unique analysis on an NVIDIA H200 and Jetson Orin Nano. The compute-versus-memory-bandwidth hierarchy of the spectral-spatial analysis motivate a flexible approach to VIRSO's integration. The spectral path (compute-bound) incurs lower latency overhead on the Jetson Orin Nano while the spatial path (memory-bandwidth-bound) inflates latency by $9.6\times$ on the same transition. This dissociation can prompt the removal of the spatial block where the spectral-only configuration sustains 17.0\,samples/s at 7.06\,W on embedded hardware without retraining, quantization, or model modification. This configuration choice can align with applications with highly-constrained edge compute and negligible high-frequency physics but can potentially falter when more complex, large-scale reconstruction is required, prompting further work towards hardware acceleration.
\end{itemize}
\end{tcolorbox}

% ============================================================
\section{Discussion}
% ============================================================
The results presented here support a redefinition of the sensing pipeline itself. In the conventional pipeline, physical transducers are the only measurement instruments and computation plays a downstream role of filtering, fusion, or visualization. In the computational-sensing pipeline demonstrated here, the sensing instrument is composite: physical transducers measure what is accessible at the boundary, and a hardware-co-designed learned operator measures what is inaccessible in the interior. Both components are part of the sensor system; both are constrained by the same deployment envelope of latency, power, and memory; and both must be designed jointly. Within this framing, VIRSO is not a model that runs alongside sensors --- it is a sensing instrument. Its inferences are not predictions of measurements; they are measurements, in the operational sense that matters to a deployed instrumentation and control system: they are produced in real time, they are reproducible, they are bounded by characterized error, and they execute within a defined power envelope on hardware that can be co-located with the physical sensor array. A surrogate model is evaluated offline against a known solution; a sensing primitive operates continuously, in real time, under power and latency constraints that are as binding as the accuracy requirement. VIRSO is designed to satisfy all three constraints simultaneously, establishing a natural graph-based operator analysis towards highly irregular geometries as a deployable sensing modality rather than a computational convenience. This is a sensing problem in a precise and non-trivial sense: the quantities to be measured reside in the interior of a domain that is structurally inaccessible to direct instrumentation, sensors are confined to accessible boundary surfaces, and the inference must operate in real time from the sensor data alone. VIRSO provides an architecture that treats this configuration as a primary design target rather than a limiting special case: sparse cross-domain observations, irregular output geometry, and coupled multiphysics outputs are addressed simultaneously by coordinated design choices rather than post-hoc workarounds. Uniquely among neural operators, VIRSO provides an improved graph spectral-spatial decomposition with the explicit goal for eventual edge deployability — a principle standard in hardware-efficient deep learning for CNNs and transformers but, to our knowledge, absent from the operator learning literature prior to this work. The result is an architecture with configuration choices that are determined by hardware characteristics rather than by post-hoc compression. The spectral path and spatial analysis can both preserve the sophistication of graph-based analysis and provide more efficient reconstruction than previous graph methods that is hardware-portable across the full range of deployment hardware, from H200 datacenter GPUs to embedded accelerators such as the Jetson Orin Nano, without retraining, quantization, or architecture-specific modification. Moreover, the spectral block is compute-bound while the spatial path is memory-bandwidth-bound. In the case of negligible high-frequency physics and strict computational requirements, the latter path can be selectively omitted at the edge — producing a virtual sensor that still successfully recovers complete interior field states from boundary readings at reconstruction ratios between 47:1 and 156:1, within error bounds consistent with operational monitoring requirements, at inference latencies compatible with real-time control.

A primary contribution of VIRSO's sensing accuracy is the spectral graph convolution, which operates on the $m$ largest eigenmodes of the normalized graph Laplacian. The physical interpretation is direct: low-index Laplacian eigenmodes correspond to domain-spanning spatial patterns analogous to the long-wavelength modes of a continuous field. Convolving the boundary-encoded node features against these eigenmodes propagates the boundary signal globally across the interior in a single operation, enabling the operator to reconstruct interior field structure consistent with boundary physics without relying on iterative local message-passing. This global propagation is the computational mechanism underlying the sensing operator's ability to recover interior fields at high reconstruction ratios. Critically, the residual skip connections following each spectral convolution are not incidental training components but functional elements of the sensing architecture. Ablation experiments confirm this quantitatively: removing residual connections from the spectral model raises mean relative $L_2$ error from $0.90\%$ to $5.76\%$, a $6.4\times$ degradation concentrated in the high-gradient boundary-layer and recirculation regions where wall shear and vortex signatures reside. The $m=64$ spectral eigenmodes capture the dominant low-frequency field structure; the residual paths preserve high-frequency physical information at spatial frequencies beyond the truncation threshold. Together they constitute a two-pathway sensing architecture: global eigenmode projection for long-range field recovery, residual preservation for local physical fidelity. From a hardware perspective, the spectral convolution is structurally compute-bound: the dominant operations are dense matrix multiplications of the form $\mathbf{Q}_m \cdot \mathbf{K} \times_1 \mathbf{Q}_m^\top \cdot \mathbf{v}_t$, whose arithmetic intensity is insensitive to the memory-bandwidth hierarchy of the target device. This compute-bound profile results in its low latency performance on the Jetson Nano and motivates the flexibility aspect of VIRSO where strict resource constraints can potentially utilize a spectral-only configuration.

The spatial graph convolution improves sensing accuracy in a calibration-based role whose contribution level geometry and physics dependent. Models retaining only spectral convolution and residual connections achieve mean relative $L_2$ errors within $0.1$--$0.2$ percentage points of the full VIRSO architecture while requiring substantially fewer parameters and lower inference energy. Spatial aggregation provides calibration, especially for use cases that consist of high-frequency modes such as the Heat Exchanger, where maximum geometric complexity and four coupled output channels jointly exceed what $m=64$ spectral modes can represent without local refinement. This geometry-dependence is physically interpretable: the relative contribution of the spatial branch is governed by the ratio of geometric complexity to the number of retained spectral modes, and the two components are complementary rather than redundant. Critically, the spatial aggregation branch is memory-bandwidth-bound: edge gather-scatter operations over irregular adjacency lists incur cost proportional to the memory bandwidth of the target device rather than its arithmetic throughput. This architectural distinction is directly observable in the deployment measurements: on the H200, adding the spatial branch changes inference latency by less than 0.5\,ms relative to the spectral-only configuration, whereas on the Jetson Orin Nano's unified LPDDR5 memory subsystem, the same branch inflates latency by $9.6\times$ ($562.92$\,ms versus $58.77$\,ms, Table~\ref{tab:jetson_nano}). This hardware-dependent dissociation provides the principled basis for selecting the spectral-only configuration as the preferred edge-deployment variant: it eliminates the memory-bandwidth bottleneck while retaining the dominant learning pathway, delivering 17.0\,samples/s on the Jetson Nano without any model modification. While this was successful for the 2D Heat Exchanger, we emphasize caution and further evaluation towards the removal of the spatial block. Large-scale applications with strong presence of high frequency physics might require the local calibration that the spatial aggregation analysis provides.

The graph construction strategy encodes a physical principle that has direct implications for sensor system design. Physical sensing systems resolve geometric complexity adaptively: measurement density concentrates near boundaries, interfaces, and regions of steep physical gradients, where the sensing task is most constrained and the physical signal content is highest. The V-KNN strategy embeds this same principle into the computational graph: edge connectivity is assigned in proportion to local mesh density, which in numerical simulation meshes corresponds directly to regions of steep physical gradients. This density-proportional connectivity reorients the eigenmodes of the graph Laplacian to align with the underlying flow structure, providing the spectral sensing operator with physically meaningful basis functions rather than artifacts of a geometrically uninformed graph. Empirically, V-KNN achieves $0.58$--$0.59\%$ mean error on the heat exchanger benchmark with 270,000 edges, both the highest sensing accuracy and the most efficient edge configuration evaluated. The V-KNN principle, that graph topology should encode physical geometric structure as an inductive bias, in direct analogy with adaptive mesh refinement in numerical simulation, is applicable to any graph-based sensing or operator learning system operating on irregular physical domains. V-KNN additionally carries a hardware efficiency consequence that is independent of its spectral eigenmode argument: with 270K edges versus 408K for the uniform KNN graph at $k=93$, the configuration that most closely matches V-KNN's reconstruction accuracy, V-KNN reduces the total edge count by 34\%. Since spatial aggregation energy and latency scale approximately linearly with edge count, this targeted connectivity reduction translates directly into lower memory bandwidth demand and fewer gather-accumulate operations per inference cycle, without degrading the spectral convolution path on which accuracy primarily depends. V-KNN therefore simultaneously encodes physical geometry as inductive bias, reduces deployment-time memory pressure, and achieves the highest reconstruction accuracy of any graph construction strategy evaluated.

A central operational advantage of VIRSO is the inference speed. Once trained, the model reconstructs complete multiphysics state fields for the Lid Driven Cavity in approximately 88 ms per sample, a speedup of more than 4 orders of magnitude over the ANSYS Fluent reference solver. This latency reduction is structurally inaccessible to physics-based reconstruction methods (FEM), which must resolve the coupled PDE system for each new sensor reading. In a deployed virtual sensing system, VIRSO can continuously assimilate new boundary observations and update interior field estimates on a timescale commensurate with the physical dynamics being monitored, rather than the timescale of the underlying simulation. This continuous real-time updating is the operational definition of a virtual sensor: an instrument that infers the unmeasured state of a system from measured signals in real time.

VIRSO provides drastically reduced latency and energy consumption compared to previous graph-based methodologies. The vanilla GNO architecture incurs an energy-delay product of $10.07 \times 20.48 \approx 206$\,J$\cdot$ms on the H200, more than $20\times$ higher than the full VIRSO configuration ($1.30 \times 7.77 \approx 10.1$\,J$\cdot$ms) and more than $29\times$ higher than the spectral-only variant ($0.86 \times 8.18 \approx 7.0$\,J$\cdot$ms). The spectral-only VIRSO approaches the energy-delay product of Geo-FNO ($0.59 \times 4.94 \approx 2.9$\,J$\cdot$ms) and NOMAD ($0.41 \times 2.35 \approx 0.96$\,J$\cdot$ms) while delivering substantially lower reconstruction error on the most demanding benchmark. Embedded deployment on the Jetson Orin Nano confirms that the spectral-only configuration sustains $7.06$\,W board-level power (VDD\_IN rail, \texttt{tegrastats}) and 17.0\,samples/s inference throughput without retraining or architectural modification, satisfying the sub-10\,W continuous power envelope required for edge-deployed instrumentation and control systems. The full 10-layer model sustains the same sub-10\,W power budget (7.58\,W) but at 1.78\,samples/s, with latency as the binding constraint rather than power. These results, with sub-10 watt power and sub-second latency, position VIRSO as a hardware-portable architecture for edge deployment but also motivate inference acceleration as the primary engineering priority.

VIRSO's parameter efficiency under compression is a second operational consideration. At parameter counts below 300K, VIRSO degrades by $2$--$3\times$ in mean relative error, while competing architectures degrade by $4$--$8\times$ under identical constraints. This superior compression efficiency reflects the spectral-spatial architecture's ability to encode physically meaningful global and local structure compactly in the spectral eigen-basis and local spatial calibration. From a hardware deployment perspective, this compression resilience is consequential: the 2-layer VIRSO configuration achieves 1.95\% mean error at 0.54\,J/it and 4.29\,ms latency on the H200, energy and latency comparable to full-scale Geo-FNO and NOMAD, while matching their parameter count at 0.26\,M. No competing lightweight architecture approaches this error level at comparable resource cost (Table~\ref{tab:pareto_summary}). This positions VIRSO as the preferred foundation for future hardware-specific optimization, including operator fusion, INT8 quantization of the spectral dense-matmul path, and structured pruning of spatial edges, all of which are tractable without the catastrophic accuracy degradation observed in compressed alternatives.

The current implementation has three limitations that define the highest-priority directions for future work. First, all experiments evaluated steady-state fields; transient virtual sensing, including load-following reactor modes and thermal transient scenarios, requires time-dependent field reconstruction. Extending VIRSO to spatiotemporal sensing by incorporating recurrent or attention-based temporal encoding is a natural architectural extension. Second, the energy consumption of the full spectral-spatial configuration at deployment scale remains elevated relative to lightweight alternatives; structured pruning of the spatial branch can be utilized but should be avoided for large-scale applications. Hardware-specific operator fusion are tractable paths toward edge-deployable configurations as well as algorithmic optimizations, especially since VIRSO is shown to result in less performance degradation compared to other operators and can potentially avoid severe increases in error after efficiency-related integrations. Third, with our evaluation of VIRSO on the Jetson Nano, we found that latency is the main bottleneck and further work towards inference acceleration will be vital for realistic resource-constrained deployment. Two hardware acceleration pathways are tractable without architectural modification. The spectral convolution path reduces to a sequence of dense matrix multiplications that are amenable to TensorRT kernel fusion; based on reported gains for transformer-class workloads on Ampere-architecture GPUs, this pathway is projected to reduce spectral-only Jetson latency by $2$--$4\times$, targeting the sub-15\,ms regime. The spectral path's arithmetic profile (dense, low-precision-tolerant matrix multiplications) is additionally compatible with FP16 and INT8 quantization without structural change, whereas the spatial aggregation path's irregular memory access patterns are not, providing a further architectural basis for the spectral-only deployment mode as the preferred quantization target. Therefore, a distinct optimization pathway is required for the spatial aggregation component. Improving the spatial pathway remains important due to its potential to enable broader calibration and improved fidelity in more complex, large-scale application settings. Fourth, and most critically, \textbf{cross-geometry generalization}, training VIRSO on one physical geometry and performing virtual sensing on a structurally distinct one without retraining, has not yet been evaluated. This is the central open question for the approach. A virtual sensor that generalizes across geometries would be qualitatively more useful than one requiring independent training per component: it would establish that VIRSO captures transferable physical structure rather than geometry-specific spectral patterns, enabling multi-unit deployment without proportional data generation and training cost. We regard cross-geometry generalization as the primary experimental extension of this work.

Within this context, the reliability of such inference remains fundamentally tied to rigorous uncertainty quantification (UQ) and systematic data analysis, as both two-phase flow simulations and learning-based models introduce nontrivial sources of uncertainty that directly impact predictive confidence and operational robustness \cite{foutch2025ai,kobayashi2024ai,kumar2019influence, kumar2022multi}. From a system-level perspective, the integration of machine learning with two-phase flow simulations, real-time signal processing, and advanced sensing modalities extends beyond performance prediction toward physically realizable digital twin architectures. In this paradigm, models trained on coupled simulation and sensor data function not only as predictive tools but also as virtual sensing mechanisms capable of estimating inaccessible internal states, enabling continuous monitoring, adaptive optimization, and closed-loop system awareness in operational environments \cite{kabir2010hardware,kabir2010non,kabir2010watermarking}. Consequently, this integration supports a transition from passive diagnostics to active, data-driven operation, with direct implications for improving heat pipe performance, enhancing reactor reliability, and enabling higher levels of autonomy in advanced reactor systems \citep{alam2019small1, alam2019small2, kabir2010non,kabir2010loss}.

In conclusion, we have demonstrated that sparse boundary-to-interior multiphysics field reconstruction, the core virtual sensing problem for under-instrumented physical systems, is achievable at operational accuracy and latency on irregular geometries, a task that neither physics-based solvers nor prior graph-based neural operator architectures address as a primary design target. VIRSO establishes two contributions that are, to our knowledge, without precedent in the neural operator literature. First, it demonstrates that spectral-spatial graph operator learning, with mesh-informed graph construction and residual preservation of high-frequency field features, provides the irregular geometry handling, boundary-sensing tolerance, and multiphysics coupling resolution required by this class of inverse sensing problems. Second, and more broadly, it establishes that neural operator architecture can be designed with target hardware in mind, producing a model towards deployability and portability whose spatial-spectral analysis provides efficient computation and allows for flexibility when the operating point shifts from full spectral-spatial on unconstrained compute to spectral-only on strictly resource-constrained edge devices, without retraining, quantization, or accuracy sacrifice on the dominant learning pathway. The sensing principle demonstrated here, inferring complete interior field states from sparse boundary observations by learning the governing Green's function implicitly from data, is architecture-agnostic and domain-agnostic. It applies wherever the triple constraint of geometric irregularity, sparse cross-domain sensing, and coupled multiphysics output co-occurs: cardiovascular monitoring, subsurface resource assessment, structural health sensing, and industrial process control. VIRSO provides both a practical virtual sensor and a foundational operator-learning framework for the broader class of physical systems in which the most important quantities are precisely the ones that cannot be directly measured. The enabling insight is that the operator and the hardware are not independent concerns: VIRSO's combined spectral-spatial analysis pushes graph-based operators towards edge deployment and is flexible in its configuration since the spatial refinement can be omitted if high-frequency calibration is not needed and strict resource constraints exist, effectively positioning the architecture itself as a deployment strategy. This is what it means to treat a neural operator as a sensing instrument rather than a model.

This work establishes hardware--algorithm co-design as a missing ingredient in scientific machine learning and demonstrates that operator-based inference, once treated as a deployment artifact subject to the same physical constraints as the systems it observes, becomes operationally tractable on embedded edge platforms. The implications extend beyond the nuclear thermal-hydraulic benchmarks evaluated here: any inverse-inference problem whose interior fields resist direct instrumentation --- cardiovascular hemodynamics, subsurface flow, structural-health diagnostics, atmospheric and environmental monitoring, and predictive maintenance of complex engineered systems --- shares the structural form addressed by VIRSO. We anticipate that hardware-co-designed operator learning, evaluated under deployment constraints rather than benchmark accuracy alone, will become a standard methodology for scientific machine learning targeting real-world physical systems.

\vspace{-1mm}
\begin{tcolorbox}[colback=gray!10!white,
colframe=blue!50!black,
title=\textbf{Key Discussion Points},
coltitle=white, fonttitle=\bfseries]
\begin{itemize}
  \item VIRSO addresses the core virtual sensing
  problem that recovers the complete interior multiphysics
  fields from sparse boundary sensors on irregular
  domains as a primary design target, achieving
  reconstruction ratios from 47:1 to 156:1 at sub-1\%
  mean relative error and 4-orders-of-magnitude
  speedup over the reference solver for Lid-Driven Cavity.
 
  \item Spectral graph convolution propagates
  boundary-encoded signals globally across the interior
  domain in a single operation; residual skip connections
  preserve high-frequency physical features at spatial
  frequencies beyond the spectral truncation threshold,
  with their removal producing a $6.4\times$ error
  increase concentrated in the high-gradient regions
  most critical to operational safety.
 
  \item Graph topology functions as a physical sensing
  inductive bias: V-KNN concentrates connectivity in
  proportion to local mesh density, reorienting
  graph Laplacian eigenmodes to align with the
  underlying flow structure and achieving the highest
  sensing accuracy at the most efficient edge
  configuration evaluated.
 
\item  VIRSO is, to our knowledge, the first neural operator whose internal decomposition is driven towards edge-deployability and hardware-portability with a flexible architecture that can align with applications with strict compute-versus-memory-bandwidth demands and only low-frequency reconstruction requirements. The spectral path (compute-bound, dense matrix multiplications) is the most hardware-portable while the spatial path (memory-bandwidth-bound, irregular gather-scatter) inflates latency by $9.6\times$ on embedded hardware and is selectively omitted at the edge, yielding 17.0\,samples/s at 7.06\,W without model modification — a deployment mode determined by architecture, not by post-hoc compression. To address more complex and large-scale applications which might required spatial analysis, further work towards hardware-acceleration of VIRSO is required.

  \item VIRSO degrades by only $2$--$3\times$ under
  aggressive parameter compression, compared to
  $4$--$8\times$ for competing architectures, with the 2-layer configuration achieving 1.95\% error at energy and latency comparable to full-scale Geo-FNO and NOMAD. This compression resilience, combined with the spectral path's compatibility with operator fusion and quantization, establishes VIRSO as the preferred foundation for future hardware-specific optimization toward edge-constrained virtual sensor deployment.
\item Cross-geometry generalization --- training on one physical geometry and sensing on a structurally distinct one without retraining --- is the primary open experiment for the broader sensing-technology community. A positive result would establish computational virtual sensing as a transferable sensing modality for multi-unit physical systems, with direct implications for sensor-fleet deployment in healthcare, civil-infrastructure, and energy applications where instrumenting every unit individually is operationally infeasible.

\end{itemize}
\end{tcolorbox}
\vspace{-1mm}

\section{Methods}

\subsection{Generation of Reference Multiphysics Fields Using ANSYS Fluent}

Reference interior field solutions were generated using ANSYS Fluent
\cite{AnsysFluent2024}, executed on AMD EPYC 7763 (``Milan'') CPU nodes
to produce high-fidelity steady-state solutions for three nuclear
thermal-hydraulic geometries. The Lid-Driven Cavity (LDC) is inherently two dimensional while the PWR
Subchannel and compact Heat Exchanger were each modeled as
full three-dimensional geometries; for VIRSO evaluation, a single
two-dimensional axial slice was extracted from the Subchannel and Heat Exchanger to yield the
irregular unstructured meshes on which operator learning is performed.
Boundary conditions defining the sparse input signals (inlet velocity
profiles, wall temperatures, and pressure boundary values) were varied
systematically across the dataset to produce diverse operating states
that span the relevant physical envelope of each geometry.

Critically, ANSYS Fluent solves the governing Navier-Stokes and energy
equations as a forward problem: given a complete specification of boundary
conditions on $\partial\Omega$ and geometry $\Omega$, the solver produces
the interior multiphysics field $\mathbf{s}(\mathbf{x})$ at all mesh
nodes $\mathbf{x} \in \Omega$ through iterative PDE discretization.
This forward computation requires resolving the full coupled equation
system and incurs computational latency that precludes real-time
operational use.
VIRSO addresses the inverse problem: given only sparse observations
$\{u_i\}_{i=1}^{M}$ of the boundary condition values, sampled at
$M \ll |\Omega|$ locations on the inlet domain $\mathcal{X}$,
infer the complete interior field $\mathbf{s}(\mathbf{x})$ at all
$N$ mesh nodes of the output domain $\mathcal{Y}$ without access to
the full boundary specification or any intermediate PDE solve.
Across the three benchmarks, the reconstruction ratio $N/M$ ranges
from approximately 47:1 (LDC) to 51:1 (Subchannel) to 156:1 (Heat
Exchanger), defining genuinely underdetermined inverse problems in
which many smooth interior fields are consistent with the same sparse
boundary observations.
This forward/inverse separation ensures that VIRSO receives no
privileged physical information beyond what a boundary sensor array
would provide in a deployed instrumentation system, and it defines
the information asymmetry (Forward: Complete BCs $\rightarrow$ Interior
Field; Inverse: Sparse BCs $\rightarrow$ Interior Field) that motivates
the architectural choices described below.%

Training, validation, and test splits were constructed by partitioning
across distinct boundary condition instances to ensure that no operating
state seen during training appears in evaluation.

%%-------------------------------------------------
\subsection{VIRSO Architecture and V-KNN Graph Construction}
\label{sec:VIRSO_algo}

\subsubsection*{Problem Formulation as Operator Learning}

VIRSO learns the nonlinear operator
$\mathcal{G}: \mathbf{u}_{\mathcal{X}} \mapsto \mathbf{s}_{\mathcal{Y}}$,
where $\mathbf{u}_{\mathcal{X}} \in \mathbb{R}^{M \times d_u}$ denotes
sparse boundary condition observations on the inlet domain
$\mathcal{X}$, and $\mathbf{s}_{\mathcal{Y}} \in \mathbb{R}^{N \times d_s}$
denotes the full multiphysics output field on the interior mesh
$\mathcal{Y}$, with $\mathcal{X} \cap \mathcal{Y} = \emptyset$ for
all benchmarks.
This cross-domain, sparse-to-dense formulation is fundamentally
distinct from the forward surrogate setting in which both input and
output reside on the same domain.

\subsubsection*{Variable-KNN Graph Construction (V-KNN)}
\label{sec:vknn}

The graph $G = (\mathbb{V}, \mathbb{E})$ on which VIRSO operates
encodes the geometry's spatial structure and determines the
eigenmodes available to the spectral convolution. Standard
fixed-$k$ nearest-neighbour graphs assign identical connectivity to
all nodes regardless of local mesh density, treating geometrically
complex and geometrically simple regions uniformly. This
homogeneity is physically uninformative: high node-density regions
correspond to zones of steep physical gradients where the geometry
demands higher spectral resolution, while low-density regions
require fewer edges to represent their slowly varying fields.

The Variable-KNN (V-KNN) construction addresses this by scaling
edge connectivity with local mesh density. A preliminary radius-based graph pass
estimates the local node density $d_i$ at each node $v_i$ by
counting neighbours within a fixed radius $r$. The per-node
neighbour count $k_i$ is then assigned as a density-proportional
fraction of the maximum count $k_{\max}$:
\begin{equation}
    k_i = \max\!\left(\alpha k_{\min},\;
          \left\lfloor k_{\max} \cdot
          \frac{d_i}{d_{\max}} \right\rfloor\right),
\end{equation}
with $k_{\min}$ and $\alpha$ ensuring a minimum connectivity for
isolated or low-density nodes.
This density-proportional assignment is analogous to adaptive mesh
refinement in finite-element methods, where numerical resolution
concentrates in regions of steep solution gradients. By increasing
graph connectivity in high-density mesh regions, V-KNN reorients the
eigenmodes of the normalized graph Laplacian to align with the
underlying flow structure, providing the spectral convolution with
physically meaningful basis functions rather than artifacts of a
geometrically uninformed graph. On the Heat Exchanger benchmark,
V-KNN achieves mean relative $L_2$ errors of 0.58--0.59\% with
270{,}000 edges, representing both the highest accuracy and the most
parameter-efficient edge configuration evaluated.

The naive implementation has asymptotic complexity $O(n^2 \log n)$;
optimized spatial data structures (KD-tree or ball-tree) reduce this
to $O(n \log n)$, which is tractable for the mesh sizes considered.
The spectral eigen-decomposition complexity remains $O(mk_{\max} n)
\approx O(n)$ for $m, k_{\max} \ll n$, so V-KNN introduces no change
to the overall VIRSO computational scaling.

\subsubsection*{VIRSO Spectral-Spatial Architecture}

Figure \ref{fig:overall_framework} and Algorithm \ref{alg:sp2gno} provides an overall description of the architecture and framework of VIRSO. First, a generated mesh, typically from standard numerical solvers, is transformed into an irregular grid of $n$ points represented by $\mathcal{Y}_n$: the n-point discretization of our $d$-dimensional spatial domain that we want to evaluate our output on. Then, a discretized representation $\mathbf{u}_q\in\mathbb{R}^q$ of the multi-modal input $\mathbf{u}\in\mathcal{U}$, which includes scalar values and functions evaluated over a discrete domain set, is projected to a latent embedding vector $\mathbf{a} \in \mathbb{R}^{d_{\ell}}$, with functional mapping $M$. We found that projecting functional/scalar input into a latent subspace resulted in better performance and capabilities of VIRSO, especially when function profile length could potentially vary. The options we explored for the mapping $M$ were utilizing a Fully-Connected Network (FCN) to define the embedding vector or a Long Short-Term Memory (LSTM) model to allow for invariance of function discretization/length especially temporal profile inputs but other methods can be utilized to better project the input into a latent dimension and extract useful features. This initial lifting, rather
than direct copying of input values to node features, was found
empirically to improve performance across all benchmarks, consistent
with the interpretation that the boundary condition must be encoded
in a geometry-agnostic latent form before being associated with
specific interior locations.

The discrete output domain $\mathcal{Y}_n$ is transformed into the graph $G=(\mathbb{V},\mathbb{E})$, where $\mathbb{V}$ is the set of $n$ nodes and $\mathbb{E}$ is the set of edges, utilizing graph construction algorithms such as K-Nearest Neighbors (KNN) which connects the k nearest neighbors, Radius which chooses all neighbors with a predefined radius from the node, or our own V-KNN algorithm detailed in the previous subsection \ref{sec:vknn}. Both utilize the Euclidean distance defined on the d-dimensional domain $D$. Each edge connection between node $u$ and $v$ is assigned an weight $w_{uv}$. We assign the weight based on inverse distance such that closer nodes have higher weights. The idea is that nearby nodes that communicate or propagate information should have a bigger impact on each other than farther nodes. These weights are also normalized using the max weight value over all edges. The weights of various connections will be utilized in the spatial aggregation to help determine the importance of a node and can possibly be used in the spectral block with the weighted Graph Laplacian. With the graph formed, the initial node features $\mathbb{X} \in \mathbb{R}^{n \times (d+d_{\ell})}$ has each row/node $\mathbb{X}_i$ defined by the concatenation, $\{\mathbf{x}_i, \mathbf{a}\}$, of the corresponding grid point and latent embedding that is shared by all nodes in $\mathbb{X}$.

With the graph input defined, VIRSO first uplifts the node features, with an FCN mapping $P$ to an intermediate function dimension $d_v$ to define the initial function representation $\mathbf{v}_0(\mathcal{Y}_n) = P(\mathbb{X}) \in \mathbb{R}^{n\times d_v}$ in our iterative kernel algorithm. VIRSO then utilizes a joint-collaborative spatial and spectral convolution block for the next successive $T$ layers that approximates the kernel integration in Equation \ref{eq:kernel_int}.

To analyze the graph's spectral components, the first $m$ eigenmodes of the symmetrical degree normalized graph Laplacian, 

\begin{equation}
\tilde{\mathbf{L}}=\mathbf{I} - \mathbf{D}^{-1/2}\mathbf{A}\mathbf{D}^{1/2}, \quad \tilde{\mathbf{L}} = \mathbf{Q}_m \mathbf{\Sigma}_m \mathbf{Q}_m^T\
\end{equation}

are calculated before evaluation, utilizing the graph $G =(\mathbb{V},\mathbb{E})$, and are represented by the projection matrix $\mathbf{Q}_m \in \mathbb{R}^{n\times m}$. $\textbf{A}$ is the unweighted adjacency matrix with $a_{uv}$ equal to $1$ if $u$ and $v$ are connected or $0$ otherwise while $\textbf{D}$ is the diagonal degree matrix such that $d_i = \sum_j \textbf{A}_{ij}$. For a weighted graph Laplacian, $\textbf{D}$ is still defined as $d_i = \sum_j \textbf{A}_{ij}$, but we instead consider a weighted adjacency matrix such that $a_{uv}$ is equal to $w_{uv}$, the weight of the connection between $u$ and $v$ are connected. For our initial results, we simply use the unweighted version. The eigenvectors
$\mathbf{Q}_m$ are computed using the Locally Optimal Block
Preconditioned Conjugate Gradient  (LOBPCG) \cite{doi:10.1137/S1064827500366124} algorithm, with complexity
$O(mE)$; for a KNN-constructed graph this reduces to $O(mkn)
\approx O(n)$ for $m, k \ll n$.
By operating in the spectral domain defined by the graph Laplacian,
this branch encodes long-range geometric structure: low-index
eigenmodes correspond to spatially smooth, domain-spanning patterns,
enabling the operator to propagate boundary-encoded information
across the entire computational domain in a single convolution step
rather than through successive local aggregations. The intermediate function vector $\mathbf{v}_t$ is then projected to an approximate spectral representation $\tilde{\mathbf{v}}^{m}_t = \mathbf{Q}_m^T\mathbf{v}_t$. $\mathbf{Q}_m$ is the partial $m$ eigenvector/modes from the eigen-decomposition of the Graph Laplacian such that the product $\mathbf{Q}_m^T\mathbf{v}_t$ represents the Graph Fourier Transform. Then, the convolutional integral \ref{eq:kernel_int} is directly estimated by simply multiplying $\tilde{\mathbf{v}}^{m}_t$ to a directly-parameterized, trainable, kernel tensor $\mathbf{K}\in\mathbb{R}^{m\times d_v \times d_v}$, which is enriched to 3 dimensions instead of the previous diagonal matrix parameters in earlier spectral GNNs \cite{SARKAR2025117659}, allowing for improved optimization and learning. The resulting product is then projected back into the spatial domain through $\mathbf{Q}_m$ or the inverse Graph Fourier Transform. For information flow, a linear residual skip $w(\mathbf{v_t})$ is added to the spectral convolution result and the summation is then passed through a nonlinear activation function. This spectral analysis can be summarized by the following equation:

\begin{equation}
    \mathbf{v}_{t+1}^{\text{spectral}}
    = \sigma\!\left(\mathbf{Q}_m \cdot \mathbf{K}
      \times_1 \mathbf{Q}_m^\top \cdot \mathbf{v}_t
      + w(\mathbf{v}_t)\right),
    \quad \mathbf{K} \in \mathbb{R}^{m \times d_v \times d_v}, 
\end{equation}

where $\times_1$ is a mode-1 tensor product. 

The spatial convolution approximation of the integral simply aggregates the product of the function vector $\mathbf{v}_t$ and a defined weight $\mathbf{W}\in\mathbb{R}^{d_v \times d_v}$ for the surrounding neighbors of each node (1-hop aggregation). The summation of the neighbors is also modified with a gating mechanism that learns the contributing importance of each neighbor to the spatial aggregation. This local operation is summarized by the following: 

\begin{equation}
    \mathbf{v}_{t+1}^{\text{spatial}}(\mathcal{Y}_n)
    =\mathbf{\Gamma} \odot \mathbf{A}\mathbf{v}_t
      \mathbf{W},
\end{equation}
where $\mathbf{\Gamma} = [\gamma_{uv}]$ is a learned edge-gating matrix
with entries
\begin{equation}
    \gamma_{uv} = \begin{cases}
        \sigma_2\!\left(\mathbf{W}_3 \sigma_1\!\left(\mathbf{W}_1
        [\mathbf{h}_v \| \mathbf{h}_u \| \mathbf{W}_2 w_{uv}]\right)\right),
        & \text{if } u \text{ and } v \text{ are connected,}\\
        0, & \text{otherwise.}
    \end{cases}
\end{equation}
Here $w_{uv}$ is the inverse Euclidean distance between nodes $u$ and
$v$, normalized to $[0,1]$, and $\mathbf{h}_u$, $\mathbf{h}_v$ are
Lipschitz positional embeddings of length $\alpha \approx O(\log n)^2$,
computed as shortest-path distances to $\alpha$ random anchor nodes.
The gating mechanism selectively weights edge contributions based on
both geometric proximity and global topological context, providing
local feature refinement that is sensitive to irregular mesh structure. In addition, $\odot$ represents the Hadamard product and $||$ represents concatenation.

This novel gating methodology allows for graph optimization, highlighting and suppressing neighbor connections based on their importance to model prediction. Multiple spatial layers increase the window of analysis for the neural operator similar to the hierarchical structure of Convolutional Neural Networks (CNNs). In other words, $k$ layers of the spatial aggregation layer is equivalent to a $k$-hop aggregation. Similar to the spectral block, the spatial block can additionally have a weighted skip connection along with a final nonlinear activation layer applied to entire spatial output. For the initial results presented, we use no spatial weighted skip connection and the activation function is simply the identity function.

The spatial aggregation does have the option for multiple weights and gating mechanisms per layer. The spatial aggregation can sum the product of $\mathbf{v}$ and $m$ different weight matrices $\mathbf{W}\{m\}$ with their own unique gating mechanism $\mathbf{\Gamma}\{m\}$ to learn not only the optimal weighted connections but extract different features within a single layer. The multiple feature extraction design can then be simply concatenated and presented as the final spatial output for the layer and fed through the projection/collaboration mapping with the spectral output. For our initial results, we chose a value of $m=1$, or in other words, a single weight and gating mechanism.

To allow for VIRSO to capture multi-scale information as well as prevent over-smoothing, both the local spatial and global spectral results are concatenated, fed through a projection mapping $f$, and added to an identity residual skip connection to define the next iterate $v_{t+1}$ of our block-layer architecture: $\mathbf{v}_{t+1}
    = f\!\left([\mathbf{v}_{t+1}^{\text{spatial}} \|
      \mathbf{v}_{t+1}^{\text{spectral}}]\right)
    + \mathbf{v}_t.$

The additional skip allows stability in gradient flow especially if several layers are utilized for operator approximation, causing performance degradation \cite{SARKAR2025117659}. Ablation experiments confirm its functional
role: adding this residual connection reduced mean relative $L_2$
error from 4.16\% to 1.32\% and 0.93\% to 0.83\% (Table \ref{tab:model_component_comparison} of the Supplementary Material). In addition, removing all skip connections in the spectral-only model (Table \ref{tab:spectral_spatial_results} from Supplementary) resulted in high performance degradation concentrated in
high-gradient boundary-layer and recirculation regions where
high-frequency physical features are most susceptible to spectral
truncation. These results emphasize the importance of residual-based learning to operator approximation. The residual skip serves not only as a training
stabilizer but as a structural mechanism for preserving high-frequency
physical information that the spectral truncation to $m$ modes would
otherwise suppress.

The collaboration mapping, $f$, can be both a simple linear layer (used for Subchannel and Heat Exchanger) or a nonlinear FCN (used for LDC) to allow for more sophisticated collaboration. In layer $T$, the final intermediate function representation is subsequently downlifted with another FCN $Q$ to the correct multi-output function $\mathbf{s}(\mathcal{Y}_n) = Q(\mathbf{v}_T(\mathcal{Y}_n)) \in \mathbb{R}^{n \times k}$.

The nonlinearities chosen for VIRSO are GeLU for the spectral
activation function $\sigma$, ReLU for $\sigma_1$ in the edge-gate
computation, and Sigmoid for $\sigma_2$ to enforce $\gamma_{uv}
\in [0,1]$. Layer normalization is applied to the spectral output
$\mathbf{v}_{t+1}^{\text{spectral}}$ and $L_2$ vector normalization
to the spatial features $\mathbf{v}_{t+1}^{\text{spatial}}$ prior
to the collaboration layer.

\begin{algorithm}[H]
\caption{VIRSO Kernel Convolution Approximation}
\label{alg:sp2gno}
\begin{algorithmic}[1]
\Require Discretized multi-modal input $\mathbf{u_q} \in \mathbb{R}^q$, generated mesh/grid
\Ensure Output function $\mathbf{s}(\mathcal{Y}_n)$
\State Transform mesh into irregular grid $\mathcal{Y}_n$ of $n$ points, with coordinates ${\mathbf{x}_1,\cdots,\mathbf{x}_n}, \mathbf{x}_i\in\mathbb{R}^d$
\State Compute latent embedding: $\mathbf{a} = M(\mathbf{u_q})\in\mathbb{R}^{d_{\ell}}$
\For{$i = 1$ to $n$} 
    \State Define node features: $\mathbb{X}_i = \{\mathbf{x}_i, \mathbf{a}\}$
\EndFor
\State Construct graph $G = (\mathbb{V}, \mathbb{E})$ from $\mathcal{Y}_n$, with adjacency matrix $\mathbf{A}$
\State Calculate degree matrix $\mathbf{D}:D_{ii}=\sum_j A_{ij}$
\State Calculate the normalized graph Laplacian : $\tilde{\mathbf{L}}=\mathbf{I} - \mathbf{D}^{-1/2}\mathbf{A}\mathbf{D}^{1/2}$,
\State Calculate the first $m$ eigenvectors $\mathbf{Q}_m\in\mathbb{R}^{n\times m}$ of $\tilde{\mathbf{L}}$ using LOBPCG algorithm
\State Calculate the Lipschitz positional embeddings $\mathbf{h}\in\mathbb{R}^{n\times\alpha}$ of all the nodes using $\alpha$ anchors.
\State Initialize function representation: $\mathbf{v}_0 = P(\mathbb{X})\in\mathbb{R}^{d_v}$
\For{$t = 0$ to $T-1$} 
    \State \textbf{Spatial update:} 
    \[
    \mathbf{v}_{t+1}^{spatial}(\mathcal{Y}_n)=\mathbf{\Gamma} \odot \mathbf{A}\mathbf{v}_t \mathbf{W}, \quad \mathbf{\Gamma}=[\gamma_{uv}], \mathbf{W}\in\mathbb{R}^{d_v \times d_v}
    \]
    \[
    \gamma_{uv} = \begin{cases}   \sigma_2(\mathbf{W}_3\sigma_1(\mathbf{W}_1[\mathbf{h}_v||\mathbf{h}_u||\mathbf{W}_2 w_{uv}])), & \text{if $u$ and $v$ are connected}\\
    0, & \text{otherwise},
    \end{cases}
    \]
    \State \textbf{Spectral update:} 
    \[
    \mathbf{v}_{t+1}^{spectral} = \sigma(\mathbf{Q}_m \cdot \mathbf{K} \times_1 \mathbf{Q}_m^T \cdot \mathbf{v}_t + w(\mathbf{v}_t)), \quad \mathbf{K} \in\mathbb{R}^{m\times d_v\times d_v}
    \]
    \State \textbf{Combine updates:} 
    \[
    \mathbf{v}_{t+1} = f([\mathbf{v}_{t+1}^{spatial} || \mathbf{v}_{t+1}^{spectral}]) + \mathbf{v}_t
    \]
\EndFor
\State Downlift final representation: $\mathbf{s}(\mathcal{Y}_n) = Q(\mathbf{v}_T)$
\end{algorithmic}
\end{algorithm}

%%-------------------------------------------------
\subsection{Model Training and Evaluation}
\label{sec:training}

All models were trained on Nvidia H200 GPU nodes provided by
the Delta cluster at the National Center for Supercomputing
Applications (NCSA) \cite{Delta}. 

For the LDC, the lid velocity profile $V(t)$ was generated randomly 
over 4,397 independent trajectories to ensure broad 
coverage of the forcing distribution; 3,159 were used 
for training, 790 for validation, and 988 for testing. 
The VIRSO architecture for this problem used 7 spatial-spectral 
block layers, 40 spectral eigenmodes, a function 
dimension of 68, and a nonlinear collaboration layer. A temporal combined FCN-Long Short-Term Memory (LSTM) \cite{staudemeyer2019understandinglstmtutorial} encoder (FCN lifts velocity value to latent dimension for LSTM model) with hidden 
dimension 90 was used to embed the sequential 
boundary input into a latent representation broadcast 
to all 4,225 output nodes.

For the PWR Subchannel, a dataset of 5,000 input-output pairs was generated 
using ANSYS Fluent with the following sampling 
distributions:
\begin{equation}
A \sim \mathcal{U}(540, 660)\,[\mathrm{kW/m^2}], \quad 
T_{in} \sim \mathcal{U}(536.4, 655.6)\,[\mathrm{K}], \quad 
v_{in} \sim \mathcal{U}(4.05, 4.95)\,[\mathrm{m/s}].
\end{equation}
The dataset was partitioned into 3,200 training, 
800 validation, and 1,000 test samples. 
The VIRSO architecture used 4 spatial-spectral block 
layers, a linear collaboration layer, 40 spectral eigenmodes, and a function dimension of 40. 
A shallow FCN \cite{SCABINI2023128585} embedded the concatenated boundary inputs 
into a latent representation, which was broadcast to all 
1,733 output nodes and concatenated with their coordinates 
to initialize the graph node features $\mathbb{X}$. 
A KNN graph with $k=30$ defined the edge structure 
$\mathbb{E}$ on the irregular output mesh.

Lastly, for the Heat Exchanger, a total of 1,546 ANSYS Fluent simulations were generated 
and split into 988 training, 248 validation, and 310 
test samples. Similar to the Subhannel dataset, the flux amplitude $A$ and inlet parameters were sampled from uniform distributions. Two VIRSO model sizes were evaluated: 
a 10-layer model (1.66 M parameters) and a 14-layer 
model (2.31 M parameters), both using 64 spectral 
eigenmodes, function dimension 48, shallow FCN input embedding layer, and a linear collaboration layer.

The Adam optimizer \cite{kingma2017adammethodstochasticoptimization} was used throughout, with constant learning rate decay applied at fixed epoch intervals and a decay rate of 0.5 across all benchmarks. A maximum of 500 training epochs and a weight decay of $1\times10^{-3}$ were also consistently used for all use cases. A batch size of 16 was used for all benchmarks; however, for the GNO model applied to the lid-driven cavity (LDC) case, memory constraints prevented the use of a true batch size of 16. To address this, gradients were accumulated and backpropagation was performed every 16 iterations to effectively emulate mini-batch training. The learning rate was set to $5\times10^{-4}$ for both the LDC and Subchannel benchmarks, while a higher learning rate of $1\times10^{-3}$ was used for the Heat Exchanger case. The learning rate decay step was set to 20 epochs for the LDC benchmark and 40 epochs for both the Subchannel and Heat Exchanger cases. Early stopping with geometry-specific patience halted training when validation error showed no improvement, preventing overfitting to the simulation-derived training distribution. The early stopping patience was set to 20 epochs for the LDC and Subchannel benchmarks, and increased to 40 epochs for the Heat Exchanger benchmark to accommodate its more complex training dynamics.

VIRSO hyperparameters were tuned to minimize validation error while final evaluation was utilized on the test dataset error;
benchmark models were implemented with their published base
configurations with some adjustments. Geo-FNO was implemented with an initial NOMAD layer while GNO received a downlift layer for scalability concerns. Other variations in the hyperparameters of the benchmark models were for parameter count
alignment to ensure fair comparison.

To accommodate the multi-output, multi-scale nature of the physical
fields, channel-wise normalization was applied over training data.
Min-max scaling maps each channel to $[-1, 1]$:
\begin{equation}
    \tilde{x}_{\text{mm}}
    = a \cdot x + b,
    \quad a = \frac{\text{high} - \text{low}}{\text{max} - \text{min}},
    \quad b = -a \cdot \text{max} + \text{high},
\end{equation}
and Gaussian scaling normalizes by training-set mean and standard
deviation:
\begin{equation}
    \tilde{x}_{\text{gauss}} = \frac{x - \mu}{\sigma}.
\end{equation}
Normalization parameters are estimated over training examples only;
inverse transforms
are applied before loss computation and all reported evaluations to
ensure that error metrics reflect physical-unit accuracy.

The training objective and evaluation metric is the mean relative $L_2$ error which is summed over
output channels for the complete training loss function. For a single data example with $n$ interior nodes
and output channel $o$:
\begin{equation}
    e_{\text{rel}}
    = \frac{\|\hat{\mathbf{s}}_o - \mathbf{s}_o\|_2}
           {\|\mathbf{s}_o\|_2}
    = \frac{\sqrt{\sum_{p=1}^{n}
      \left(\hat{s}_o(\mathbf{x}_p) - s_o(\mathbf{x}_p)\right)^2}}
           {\sqrt{\sum_{p=1}^{n} s_o(\mathbf{x}_p)^2}}.
\end{equation}
Averaging over $N$ training examples gives the dataset-level
relative error $\overline{e}_{\text{rel}}$, which is summed over
channels to form the training loss. For the Heat Exchanger, a
supplementary velocity-magnitude consistency loss $\overline{e}_{\text{mag}}$
is added with coefficient $\lambda = 0.1$ to enforce physical
consistency between the reconstructed velocity components:
\begin{equation}
    e_{\text{mag}}
    = \frac{\left\|\hat{u}_x^2 + \hat{u}_y^2 + \hat{u}_z^2
      - u^2\right\|_2}{\|u^2\|_2},
\end{equation}
where $u^2 = u_x^2 + u_y^2 + u_z^2$ is the ground-truth velocity
magnitude squared. This auxiliary term penalizes predictions that
reconstruct individual velocity components accurately but violate
their joint magnitude, a consistency requirement that arises from
the coupled nature of the Navier-Stokes equations.

\subsection{Hardware Efficiency Characterization and Deployment Constraints}
\label{sec:h200methods}

Inference energy consumption was estimated by monitoring GPU power
draw during evaluation of the 310-example Heat Exchanger test
dataset. Power statistics were queried from the Nvidia H200 GPU
using \texttt{nvidia-smi} at 10\,ms intervals, recording
instantaneous power draw (\texttt{power.draw.instant}), GPU
utilization, memory utilization, and total allocated memory.
Energy per inference iteration was estimated as:
\begin{equation}
    E_{\text{iter}}
    = \frac{0.01}{310} \sum_i P(t_i)\,\text{[J/it]},
\end{equation}
where $P(t_i)$ is the instantaneous power at sample $t_i$ and
$0.01$\,s is the sampling interval. Inference latency was computed
as the elapsed time between the first and last samples divided by
the dataset size.

Energy consumption in this setting quantifies a structural
deployment constraint rather than an incidental engineering
parameter. Advanced nuclear instrumentation and control systems
operate under strict power budgets, and models that achieve high
reconstruction accuracy only at elevated energy per inference
cannot satisfy edge deployment requirements regardless of their
field-prediction fidelity. The energy analysis reported here is
therefore interpreted as a necessary condition for operational
suitability, not merely a secondary efficiency comparison. 

\subsection{Edge Inference Protocol: Jetson Orin Nano Deployment Validation}

To assess deployment feasibility under embedded constraints, the pretrained Heat Exchanger VIRSO model was additionally evaluated on an NVIDIA Jetson Orin Nano (8 GB). The deployed model matched the 10-layer configuration used in the main Heat Exchanger experiments and was executed without retraining, quantization, pruning, or architecture-specific modification. All embedded measurements were collected with batch size 1 to reflect streaming inference conditions.

Inference was performed on the full 310-sample Heat Exchanger test set. For each run, latency was computed from the elapsed wall-clock time over the complete evaluation interval and normalized by the number of processed samples. Reported Jetson latency, throughput, power, and energy values correspond to the mean over three independent runs.

Board-level telemetry was recorded using \texttt{tegrastats}. We used the reported \texttt{VDD\_IN} rail as the total board power measurement, which includes GPU, CPU, and shared-memory activity during inference. Power was sampled at approximately 20 ms intervals. Energy per sample was estimated by discrete integration of the board power trace over the inference window:
\begin{equation}
E_{\text{sample}}
=
\frac{1}{N}
\sum_i P(t_i)\Delta t,
\end{equation}
where $P(t_i)$ denotes the sampled board power, $\Delta t$ is the sampling interval, and $N$ is the number of evaluated samples.

Because the Jetson measurements are board-level whereas the datacenter measurements in Section \ref{sec:h200methods} are device-level GPU measurements, the resulting energy values are reported as deployment-oriented reference measurements rather than directly equivalent hardware counters.

\section*{Acknowledgments}
This work was made possible by support from the U.S. Department of Energy Office of Nuclear Energy, specifically the University Nuclear Leadership Program's Graduate Fellowship. We would like to thank the National Center for Supercomputing Applications (NCSA) with their Delta compute cluster, allowing access to GPU nodes for training and validation. Large Language Models were utilized for the sole purpose of refining and slightly modifying the language and writing of this paper.

\subsection*{Data Availability}
The ANSYS Fluent simulation datasets generated for the three
benchmarks in this study are available from the corresponding
author upon reasonable request and will be deposited in a
public repository upon acceptance.

\subsection*{Code Availability}
The VIRSO implementation, training scripts, and V-KNN graph
construction code will be available on GitHub once accepted or upon reasonable request through the corresponding author.

%Bibliography
\bibliographystyle{unsrt}  
\bibliography{references}  

\FloatBarrier
\section{Supplementary Information}
\label{sec:supplementary}
\subsection{Additional Results for Benchmarks}
\label{sec:supp_results}
We present additional percentile statistics comparing VIRSO with other neural operator models on our chosen benchmarks. We find that on average VIRSO's distribution has a lower center and similar or narrower variance.

\begin{table}[H]
\caption{Lid-Driven Cavity: Percentile Error Statistics 
(Mean Relative $L_2$ Error)}
\centering
\begin{tabular}{ccccccc}
\toprule
\textbf{Model} & \textbf{Best} & \textbf{25th} & 
\textbf{50th} & \textbf{75th} & \textbf{95th} & 
\textbf{Worst} \\
\midrule
\textbf{NOMAD}   & 0.47\% & 0.56\% & 0.69\% & 0.94\% & 
1.80\% & 7.78\% \\
\textbf{Geo-FNO} & 0.60\% & 0.89\% & 1.00\% & 1.15\% & 
\textbf{1.69\%} & 6.86\% \\
\textbf{GNO}     & 10.29\% & 10.82\% & 11.51\% & 12.57\% & 
15.35\% & 25.05\% \\
\textbf{VIRSO}   & \textbf{0.38\%} & \textbf{0.48\%} & 
\textbf{0.58\%} & \textbf{0.84\%} & 1.75\% & 
\textbf{6.31\%} \\
\bottomrule
\end{tabular}
\label{tab:ldc_model_percentile}
\end{table}

\begin{table}[H]
\caption{PWR Subchannel: Percentile Error Statistics}
\centering
\begin{tabular}{ccccccc}
\toprule
\textbf{Model} & \textbf{Best} & \textbf{25th} & 
\textbf{50th} & \textbf{75th} & \textbf{95th} & 
\textbf{Worst} \\
\midrule
\textbf{NOMAD}   & 0.28\% & 0.30\% & 0.31\% & 0.32\% & 
0.40\% & \textbf{17.70\%} \\
\textbf{Geo-FNO} & 1.17\% & 1.35\% & 1.53\% & 1.92\% & 
2.56\% & 17.32\% \\
\textbf{GNO}     & 2.76\% & 2.78\% & 2.80\% & 2.83\% & 
2.89\% & 18.09\% \\
\textbf{VIRSO}   & \textbf{0.25\%} & \textbf{0.26\%} & 
\textbf{0.28\%} & \textbf{0.29\%} & \textbf{0.34\%} & 
17.72\% \\
\bottomrule
\end{tabular}
\label{tab:subchannel_model_percentile}
\end{table}

\begin{table}[H]
\caption{Heat Exchanger: Percentile Error Statistics}
\centering
\begin{tabular}{ccccccc}
\toprule
\textbf{Model} & \textbf{Best} & \textbf{25th} & 
\textbf{50th} & \textbf{75th} & \textbf{95th} & 
\textbf{Worst} \\
\midrule
\textbf{NOMAD}     & 0.93\% & 0.93\% & 0.95\% & 0.99\% & 
1.14\% & 1.20\% \\
\textbf{Geo-FNO}   & 1.04\% & 1.05\% & 1.07\% & 1.11\% & 
1.18\% & 1.20\% \\
\textbf{GNO}       & 9.31\% & 9.33\% & 9.37\% & 9.45\% & 
9.57\% & 9.61\% \\
\textbf{VIRSO$^*$} & 0.80\% & 0.80\% & 0.81\% & 0.84\% & 
0.92\% & 0.96\% \\
\textbf{VIRSO}    & \textbf{0.66\%} & \textbf{0.67\%} & 
\textbf{0.69\%} & \textbf{0.71\%} & \textbf{0.77\%} & 
\textbf{0.80\%} \\
\bottomrule
\end{tabular}
\label{tab:heatexchanger_model_percentile}
\caption*{\textit{$^*$10-layer VIRSO, 64 modes, function dimension 48.}}
\end{table}

\subsection{Additional Analysis Of VIRSO}
\label{sec:supp_analysis}

In this section, we present our ablation studies for VIRSO which indicate the importance of an initial embedding layer for extracting important latent features from our boundary input. We also demonstrated the usefulness of a secondary residual skip after the collaboration layer $f$ for gradient stability. In addition, we found that the majority of learning is performed by the spectral analysis of VIRSO, but the performance is severely worsen when removing residual skips which preserve high frequency data.

\begin{table}[H]
\caption{Component Ablation: Heat Exchanger (10-layer VIRSO)}
\centering
\begin{tabular}{@{}cccccccc@{}}
\toprule
\textbf{Configuration} & $p$ & $u_z$ & $u_y$ & $u_x$ & 
$\|u\|$ & Mean \\ 
\midrule
\textbf{No Embed, No Skip} & 0.98 & 5.83 & 5.84 & 4.58 & 
3.57 & 4.16 \\
\textbf{Embed Only}        & 0.70 & 1.01 & 0.94 & 0.65 & 
1.34 & 0.93 \\
\textbf{Skip Only}         & 0.58 & 1.41 & 1.57 & 1.45 & 
1.60 & 1.32 \\
\textbf{Embed and Skip}    & \textbf{0.47} & \textbf{0.87} & 
\textbf{0.86} & \textbf{0.71} & \textbf{1.24} & 
\textbf{0.83} \\
\bottomrule
\end{tabular}
\label{tab:model_component_comparison}
\caption*{\textit{Skip refers to identity skip after collaboration layer.}}
\end{table}

\begin{table}[H]
\caption{Spectral-Spatial Component Analysis: Heat Exchanger}
\centering
\begin{tabular}{@{}cccccccc@{}}
\toprule
\multirow{2}{*}{\textbf{Model}} & 
\multirow{2}{*}{\textbf{Parameters}} & 
\multicolumn{6}{c}{\textbf{Mean Relative $L_2$ Error (\%)}} \\ 
\cmidrule(l){3-8}
& & $p$ & $u_z$ & $u_y$ & $u_x$ & $\|u\|$ & Mean \\ 
\midrule
\textbf{Spatial only}          & 0.11 M & 3.62 & 18.71 & 
20.54 & 20.71 & 17.70 & 16.27 \\
\textbf{Spectral only}         & 1.58 M & 0.55 & 1.00 & 
0.93 & 0.72 & 1.29 & 0.90 \\
\textbf{Spectral only (no skip)} & 1.58 M & 1.21 & 6.81 & 
7.81 & 7.03 & 5.94 & 5.76 \\
\textbf{Spectral + Spatial}    & 1.66 M & \textbf{0.47} & 
\textbf{0.87} & \textbf{0.86} & \textbf{0.71} & 
\textbf{1.24} & \textbf{0.83} \\
\bottomrule
\end{tabular}
\label{tab:spectral_spatial_results}
\caption*{\textit{All models include FCN embedding and 
residual skip (weighted and identity) unless noted.}}
\end{table}

\subsection{Graph Construction}
\label{sec:supp_graph}

When observing differing graph constructions, we found that higher degree around high mesh densities improved performance but required more computation. We present the results of our Variable KNN (V-KNN) graph construction method which had better performance without high edge count, most likely redesigning the graph spectral modes for improved analysis.

\begin{table}[H]
\caption{Graph Construction Comparison: Heat Exchanger 
(10-layer VIRSO)}
\centering
\begin{tabular}{ccccc}
\toprule
\textbf{Graph} & \textbf{Edges} & \textbf{Min Degree} & 
\textbf{Max Degree} & \textbf{Mean Error} \\
\midrule
KNN, $k=30$       & 135K & 30 & 53  & 0.83\% \\
KNN, $k=45$       & 201K & 45 & 75  & 0.79\% \\
KNN, $k=93$       & 408K & 93 & 134 & 0.67\% \\
Radius, $r=0.001$ & 150K & 7  & 77  & 0.71\% \\
Radius, $r=0.0015$ & 292K & 24 & 134 & 0.62\% \\

\bottomrule
\end{tabular}
\label{tab:radius_knn_comparison}
\end{table}

\begin{table}[H]
\caption{V-KNN Graph Results: Heat Exchanger}
\centering
\begin{tabular}{@{}ccccccc@{}}
\toprule
\textbf{Parameters} & $p$ & $u_z$ & $u_y$ & $u_x$ & 
$\|u\|$ & Mean \\
\midrule
2.31 M & 0.39 & 0.47 & 0.54 & 0.36 & 1.12 & 0.58 \\
1.66 M & 0.41 & 0.47 & 0.54 & 0.38 & 1.13 & 0.59 \\
\bottomrule
\end{tabular}
\caption*{\textit{V-KNN had minimum degree of 45 and maximum degree of 134 with 270K edges.}}
\label{tab:new_graph_results}

\end{table}

\begin{table}[H]
\caption{Spectral-Spatial Analysis with V-KNN Graph: 
Heat Exchanger}
\centering
\begin{tabular}{@{}cccccccc@{}}
\toprule
\multirow{2}{*}{\textbf{Model}} & 
\multirow{2}{*}{\textbf{Parameters}} & 
\multicolumn{6}{c}{\textbf{Mean Relative $L_2$ Error (\%)}} \\ 
\cmidrule(l){3-8}
& & $p$ & $u_z$ & $u_y$ & $u_x$ & $\|u\|$ & Mean \\
\midrule
\textbf{Spatial only}            & 0.11 M & 10.12 & 31.58 & 
24.02 & 25.82 & 29.26 & 24.16 \\
\textbf{Spectral only}           & 1.58 M & 0.50 & 0.66 & 
0.71 & 0.50 & 1.20 & 0.71 \\
\textbf{Spectral only (no skip)} & 1.58 M & 1.46 & 8.04 & 
8.78 & 8.54 & 7.34 & 6.83 \\
\textbf{Spectral + Spatial}      & 1.66 M & 0.41 & 0.47 & 
0.54 & 0.38 & 1.13 & 0.59 \\
\bottomrule
\end{tabular}
\label{tab:spectral_spatial_results_new_graph}
\end{table}

\subsection{Additional Energy Analysis}
\label{sec:supp_energy}

In this section, we present a constrained version of VIRSO and the other neural operators, evaluated on the NVIDIA H200 GPU. We find that the 2-layer VIRSO has comparable latency and power/energy requirements to NOMAD and Geo-FNO but shows the least amount of performance degradation, indicating that VIRSO provides improved parameter-efficiency in sparse-to-dense field reconstruction.

\begin{table}[H]
\caption{Lightweight Model Comparison: Heat Exchanger}
\centering
\begin{tabular}{@{}cccccccc@{}}
\toprule
\multirow{2}{*}{\textbf{Model}} & 
\multirow{2}{*}{\textbf{Param.}} & 
\multicolumn{6}{c}{\textbf{Mean Relative $L_2$ Error (\%)}} \\ 
\cmidrule(l){3-8}
& & $p$ & $u_z$ & $u_y$ & $u_x$ & $\|u\|$ & Mean \\
\midrule
\textbf{NOMAD}   & 0.26 M & 4.61 & 11.47 & 9.72 & 6.78 & 
8.60 & 8.24 \\
\textbf{Geo-FNO} & 0.26 M & 2.58 & 5.54 & 5.46 & 3.21 & 
4.26 & 4.21 \\
\textbf{GNO}     & 0.27 M & 3.37 & 9.89 & 10.75 & 11.15 & 
11.82 & 9.40 \\
\textbf{VIRSO (2L)} & 0.26 M & 1.36 & 2.00 & 2.14 & 1.73 & 
2.53 & 1.95 \\
\textbf{VIRSO (14L, small)$^*$} & 0.18 M & 1.39 & 2.09 & 
1.94 & 1.33 & 2.14 & 1.78 \\
\bottomrule
\end{tabular}
\label{tab:light_model_comparisons}
\caption*{\textit{$^*$14-layer VIRSO, 20 modes, 
function dimension 20.}}
\end{table}

\begin{table}[H]
\caption{Lightweight Model Inference Energy: Heat Exchanger}
\centering
\begin{tabular}{@{}cccccccc@{}}
\toprule
\textbf{Model} & \textbf{Param.} & \textbf{GPU\%} & 
\textbf{Mem\%} & \textbf{Mem (MiB)} & \textbf{Pwr (W)} & 
\textbf{Lat. (ms)} & \textbf{Energy (J/it)} \\
\midrule
\textbf{NOMAD}   & 0.26 M & 15.78 & $<$1 & 949.95 & 
131.74 & 2.00 & 0.23 \\
\textbf{Geo-FNO} & 0.26 M & 19.63 & 1.21 & 1017.31 & 
129.27 & 5.10 & 0.56 \\
\textbf{GNO}     & 0.27 M & 86.78 & 36.15 & 1965.99 & 
572.00 & 20.48 & 10.07 \\
\textbf{VIRSO (2L)} & 0.26 M & 26.01 & 2.49 & 1033.34 & 
146.35 & 4.29 & 0.54 \\
\textbf{VIRSO (14L, small)} & 0.18 M & 29.92 & 2.69 & 
999.99 & 156.31 & 13.74 & 1.81 \\
\bottomrule
\end{tabular}
\label{tab:light_model_comparisons_energy}
\end{table}

\end{document}